%% file: paper.tex
\RequirePackage{fix-cm}
\documentclass[smallextended]{svjour3}       
\smartqed  
\usepackage{graphicx}
\usepackage{booktabs}
\usepackage{tabularx}
\usepackage{tablefootnote}
\usepackage{threeparttable}
\usepackage{amsmath}
\usepackage{amssymb}
\usepackage{xspace}
\usepackage[center]{subfigure} 
\usepackage[ruled,vlined,linesnumbered]{algorithm2e}
\usepackage[font=small]{caption}
\usepackage{algorithmic}
\usepackage{enumitem}
\setitemize{noitemsep,topsep=0pt,parsep=0pt,partopsep=0pt}
\usepackage{color}
\usepackage{wasysym}
\usepackage{natbib}
\usepackage[toc,page]{appendix}
\usepackage{enumitem}

\begin{document}

\title{Fast and Accurate Pseudoinverse with Sparse Matrix Reordering and Incremental Approach
}


\author{Jinhong Jung         \and
        Lee Sael 
}


\institute{Jinhong Jung \at
		   Jeonbuk National University, 
		   567 Baekje-daero, Deokjin-gu, Jeonju 54907, Republic of Korea \\
			\email{jinhongjung@jbnu.ac.kr}           
           \and
           Lee Sael (Corresponding author.)\at
           Ajou University, 206 Worldcup-ro, Yeongton-ju, Suwon 16499, Republic of Korea\\
           \email{sael@ajou.ac.kr}
}

\date{Received: date / Accepted: date}

\maketitle

\input{def}

\begin{abstract}
How can we compute the pseudoinverse of a sparse feature matrix efficiently and accurately for solving optimization problems?
A pseudoinverse is a generalization of a matrix inverse, which has been extensively utilized as a fundamental building block for solving linear systems in machine learning.
However, an approximate computation, let alone an exact computation, of pseudoinverse is very time-consuming due to its demanding time complexity, which limits it from being applied to large data.
In this paper, we propose \method (\methodlong), a novel incremental singular value decomposition (SVD) based pseudoinverse method for sparse matrices.
Based on the observation that many real-world feature matrices are sparse and highly skewed, \method reorders and divides the feature matrix and incrementally computes low-rank SVD from the divided components.
To show the efficacy of proposed \method, we apply them in real-world multi-label linear regression problems.
Through extensive experiments, we demonstrate that \method computes the pseudoinverse faster than other approximate methods without loss of accuracy.
Results imply that our method efficiently computes the low-rank pseudoinverse of a large and sparse matrix that other existing methods cannot handle with limited time and space.
\end{abstract}

\section{Introduction}
\label{sec:introduction}
\input{001introduction}

\section{Preliminaries}
\label{sec:preliminaries}
\input{002preliminaries}

\section{Proposed Method}
\label{sec:proposed_method}
\input{003proposed_method}

\section{Experiment}
\label{sec:experiment}
\input{004experiment}

\vspace{-5mm}
\section{Conclusion}
\label{sec:conclusion}
\input{005conclusion}

\bibliographystyle{spbasic}      
\bibliography{myref}




\end{document}

%% file: def.tex
\newtheorem{application}{Application}

\renewcommand{\algorithmicrequire}{\textbf{Input:}}
\renewcommand{\algorithmicensure}{\textbf{Output:}}

\newcommand{\mat}[1]{\mathbf{#1}}
\newcommand{\matt}[1]{\mathbf{#1}^{\top}}
\newcommand{\mati}[1]{\mathbf{#1}^{-1}}
\newcommand{\vect}[1]{\mathbf{#1}}
\newcommand{\vectt}[1]{\mathbf{#1}^{\top}}

\newcommand{\R}[2]{\mathbb{R}^{#1 \times #2}}
\newcommand{\U}[2]{\mat{U}_{#1 \times #2}}
\renewcommand{\S}[2]{\mat{\Sigma}_{#1 \times #2}}
\newcommand{\VT}[2]{\matt{V}_{#1 \times #2}}
\renewcommand{\O}[2]{\mat{0}_{#1 \times #2}}
\newcommand{\I}[2]{\mat{I}_{#1 \times #2}}

\newcommand{\Ut}[2]{\mat{\tilde{U}}_{#1 \times #2}}
\newcommand{\St}[2]{\mat{\tilde{\Sigma}}_{#1 \times #2}}
\newcommand{\VtT}[2]{\matt{\tilde{V}}_{#1 \times #2}}

\newcommand{\UtT}[2]{\matt{\tilde{U}}_{#1 \times #2}}
\newcommand{\pSt}[2]{\mat{\tilde{\Sigma}}^{\dagger}_{#1 \times #2}}
\newcommand{\Vt}[2]{\mat{\tilde{V}}_{#1 \times #2}}

\newcommand{\UT}[2]{\matt{U}_{#1 \times #2}}
\newcommand{\pS}[2]{\mat{\Sigma}^{\dagger}_{#1 \times #2}}
\newcommand{\V}[2]{\mat{V}_{#1 \times #2}}

\newcommand{\A}{\mat{A}}
\newcommand{\An}[1]{\A_{#1}}
\newcommand{\pinv}[1]{\mat{#1}^{\dagger}}
\newcommand{\hpinv}[1]{\mat{\hat{#1}}^{\dagger}}

\newcommand{\ltnorm}[1]{\lVert #1 \rVert_{2}}

\newcommand*{\QEDA}{\hfill\ensuremath{\blacksquare}}%
\newcommand*{\QEDB}{\hfill\ensuremath{\square}}%

\newcommand{\method}{\textsc{FastPI}\xspace}
\newcommand{\methodlong}{{Fast PseudoInverse}\xspace}

\newcommand{\vertdots}{\underset{\big{\overset{\cdot}{\cdot}}}{\cdot}}
\newcommand{\diagdots}{_{^{\big\cdot}\cdot _{\big\cdot}}}

%% file: 001introduction.tex
A pseudoinverse is a generalized inverse method for all types of matrices~\citep{ben2003generalized} that play a crucial role in obtaining best-fit solutions to the linear systems even when unique solutions do not exist~\citep{strang2006linear}.
Pseudoinverses have been studied by many researchers in various domains, including mathematics and machine learning, from the viewpoint of theory~\citep{ben2003generalized}, computational engineering~\citep{golub2012matrix}, and applications~\citep{guo2019pseudoinverse,xu2018pseudoinverse,he2016novel,spyromitros2016multi,chen2012feature,horata2013robust}.

Although pseudoinverses have been widely applied, applications were limited to small data due to their high computational complexity. 
More specifically, the most widely applied pseudoinverse is the Moore-Penrose inverse; and the most elegant and precise solution for obtaining the Moore-Penrose inverse is by utilizing a singular value decomposition (SVD).
However, calculating SVDs are impractical for large matrices, i.e., the time complexity of a full-rank SVD for an $m \times n$ matrix is $min(O(n^2m),O(nm^2))$ \citep{trefethen1997numerical}.
Low-rank approximation techniques~\citep{halko2011finding, feng2018fast} 
have been proposed to reduce the time complexity problem. However, costs can still be improved, especially for handling large matrices using larger rank approximations for higher accuracies; e.g.,
$O(mn\log(r)+(m+n)r^2)$ when a randomized algorithm is used, where $r$ is the low-rank, is still large.


In this paper, we propose \method (\methodlong), a novel approximation algorithm for computing pseudoinverse efficiently and accurately based on sparse matrix reordering and incremental low-rank SVD.
\method was motivated by the observation that many real-world feature matrices are highly sparse and skewed.
Based on this observation, \method reorders a feature matrix such that its non-zero elements are concentrated at the bottom right corner leaving a large sparse area at the top left of the feature matrix (Figure~\ref{fig:matrix_reordering}(e)).
The reordered matrix is split into four submatrices, where one of the submatrices is the large and sparse rectangular block diagonal matrix, whose SVD is easy-to-compute.
\method efficiently obtains the approximate pseudoinverse of the feature matrix by performing incremental low-rank SVD starting from the SVD of this block diagonal submatrix.
Experiments show that \method successfully approximates the pseudoinverse faster than compared methods without loss of accuracy in the multi-label linear regression problem.
%
Our contributions are the followings:
\begin{itemize}
    \item {\textbf{Observation.}
    We observed that a sparse feature matrix can be transformed to a bipartite network (Definition~\ref{definition:bipartite_network}) characteristic with a highly skewed node degree distribution (Figure~\ref{fig:degree_distributions}).
    %
    }
    \item {\textbf{Method.}
    We propose \method, a novel method for efficiently and accurately obtaining the approximate pseudoinverse with sparse matrix reordering and incremental SVD (Algorithm~\ref{alg:method}).
    }
    \item {\textbf{Experiment.}
    We show \method computes an approximate pseudoinverse faster than its competitors for most datasets without loss of accuracy in the multi-label linear regression experiments (Figures~\ref{fig:accuracy:p3}~and~\ref{fig:running_time}).
    }
\end{itemize}

%% file: 002preliminaries.tex
\setlength{\tabcolsep}{16pt}
\begin{table}
	\centering
	
	\caption{
		\label{tab:symbols}
		Table of symbols.
	}
	\begin{tabular}{cl}
		\toprule
		\textbf{Symbol} & \textbf{Definition} \\
		\midrule
		$m$ & {\small number of training instances} \\
		$n$ & {\small number of features} \\
		$L$ & {\small number of labels} \\
        $\mat{A} \in \R{m}{n}$ & {\small input feature matrix}\\
        $\pinv{A} \in \R{n}{m}$ & {\small pseudoinverse of the feature matrix}\\
        $\U{m}{r}$ & {\small $m \times r$ matrix of left singular vectors}\\
        $\S{r}{r}$ & {\small $r \times r$ diagonal matrix of singular values}\\
        $\VT{r}{n}$ & {\small $r \times n$ matrix of right singular vectors}\\
        $\alpha$ & {\small target rank ratio in Algorithm 1 where $0 < \alpha \leq 1$} \\
        $r$ & {\small target rank, i.e., $r = \lceil \alpha n \rceil$ for $m \times n$ matrix when $m > n$} \\
        $\An{ij} \in \R{m_i}{n_j}$ & {\small $(i, j)$-th submatrix of reordered $\A$} \\
        $k$ & {\small hub selection ratio in Algorithm 2 where $0 < k < 1$}\\ 
        $m_1$ \& $n_1$ & {\small number of spoke instance and feature nodes, respectively}\\
        $m_2$ \& $n_2$ & {\small number of hub instance and feature nodes, respectively}\\
        $B$ & {\small number of rectangular blocks in $\An{11}$}\\
        $m_{1i}$ \& $n_{1i}$ & {\small height and width of $i$-th block in $\An{11}$, respectively}\\
        $|\A|$ & {\small number of non-zero entries in $\A$}\\
		\bottomrule
	\end{tabular}
\end{table}

We describe the preliminaries on pseudoinverse and singular value decomposition (SVD), and provide the formal definition of the problem and target application handled in this paper.
Symbols used in the paper are summarized in Table~\ref{tab:symbols}.

\subsection{Pseudoinverse and SVD}
In many machine learning models, a training data is represented as a feature matrix
denoted by $\mat{A} \in \R{m}{n}$, where $m$ is the number of training instances and $n$ is the number of features.
Learning optimal model parameters often involves pseudoinverse $\mat{A}^{\dagger} \in \R{n}{m}$.
The Moore-Penrose inverse is the most accurate and widely used generalized matrix inverse that can be solved using SVD as follows.

\begin{problem} \textnormal{\textbf{(Solving Moore-Penrose
Inverse via low-rank SVD~\citep{golub2012matrix})}}
\label{problem:pseudoinverse}
    For feature matrix $\mat{A}\in\R{m}{n}$, let $\mat{A}$ be decomposed into $\U{m}{r}\S{r}{r}\VT{r}{n}$,
where $\U{m}{r}$ and $\V{n}{r}$ are orthogonal matrices and $\S{r}{r}$ is  diagonal with $r$ singular values.
    If $r$ is the rank of $\mat{A}$, the pseudoinverse $\pinv{A}$ of $\mat{A}$ is given by $\pinv{A} = \V{n}{r}\pS{r}{r}\UT{r}{m}$.
    Otherwise, for a given target rank $r$, it results in a best approximate pseudoinverse $\pinv{A} \approx \V{n}{r}\pS{r}{r}\UT{r}{m}$. 
\end{problem}

The state-of-the-art low-rank SVD is randomized-SVD with the computational complexity of $O(mn\log(r)+(m+n)r^2)$.
Randomized-SVD utilizes randomized algorithm with oversampling technique (see the details in Section~\ref{sec:experiment:setting}) for efficient computation~\citep{halko2011finding}.
In cases of sparse matrices, Krylov subspace-based methods have also been shown to be efficient~\citep{baglama2005augmented}. However, both methods target problems that require very small ranks, while as accurate approximations of pseudoinverses require relatively large rank approximations of SVDs. Thus, the costs of existing low-rank SVDs are still too heavy for practical applications of pseudoinverses on large feature matrices.

\subsection{Target Application of Pseudoinverse}
\label{sec:preliminaries:problem_definition}
We describe our target application, multi-label linear regression based on pseudoinverse as follows:
\begin{application}
\textnormal{\textbf{(Multi-label Linear Regression~\citep{yu2014large})}}
\label{application:mtlr}
    Given feature matrix $\mat{A}\!\in\!\R{m}{n}$ and label matrix $\mat{Y}\!\in\!\R{m}{L}$ where $m\!>\!n$, $L$ is the number of labels, and each row of $\mat{Y}$ is a binary label vector of size $L$,
the goal is to
    learn parameter $\mat{Z}\!\in\!\R{n}{L}$ satisfying $\mat{A}\mat{Z} \simeq \mat{Y}$ to estimate the score vector $\vect{\hat{y}} = \matt{Z}\vect{a}$ for a new feature vector $\vect{a} \in \mathbb{R}^{n}$.
\end{application}

The linear system for unknown $\mat{Z}$ is over-determined when $m > n$;
thus, the solution for $\mat{Z}$ is obtained by minimizing the least square error $\lVert \mat{A}\mat{Z} - \mat{Y} \rVert^{2}_{F}$, which results in the closed form solution $\mat{Z} = \pinv{A}\mat{Y}$~\citep{chen2012feature}. 
As described in Problem~\ref{problem:pseudoinverse}, $\pinv{A} \simeq \V{n}{r}\pS{r}{r}\UT{r}{m}$,
where the equality holds when $r$ is the rank of $\mat{A}$.
Hence, the SVD results can be used to compute pseudoinverse exactly or approximately in a multi-label linear regression.

%% file: 003proposed_method.tex
\setlength{\algomargin}{0.5em}
\begin{algorithm}[t!]
    \begin{algorithmic}[1]
        \caption{\method}
        \label{alg:method}
        \REQUIRE input matrix $\A \in \R{m}{n}$ and target rank ratio $0 < \alpha \leq 1$
        \ENSURE approximate pseudoinverse $\pinv{A}$
        \STATE reorder $\A$ using Algorithm~\ref{alg:matrix_reordering}, and divide $\A$ into $\begin{bmatrix}
                                \An{11} \!&\! \An{12} \\
                                \An{21} \!&\! \An{22}
                            \end{bmatrix}$\label{alg:method:reorder}
        \STATE $\U{m_1}{s}\S{s}{s}\VT{s}{n_1} \leftarrow$ compute the SVD result for $\An{11}$ with target rank $s = \lceil \alpha n_1 \rceil$ according to Equation~\eqref{eq:block_diagonal} \label{alg:method:svda11}
        \BlankLine
        \STATE $\U{m}{s}\S{s}{s}\VT{s}{n_1} \leftarrow$ incrementally update the SVD result for $\An{21}$  with target rank $s = \lceil \alpha n_1 \rceil$ according to Equation~\eqref{eq:row_incremental_update} \label{alg:method:row_inc_up}
        \STATE $\U{m}{r}\S{r}{r}\VT{r}{n} \leftarrow$ incrementally update the SVD result for $\begin{bmatrix} \An{12} \\ \An{22} \end{bmatrix}$ with target rank $r = \lceil \alpha n \rceil$ according to Equation~\eqref{eq:column_incremental_update}
            \label{alg:method:col_inc_up}
        \BlankLine
        \STATE $\pinv{A} \leftarrow $ solve pseudoinverse (Problem~\ref{problem:pseudoinverse})
        \label{alg:method:final}
        \BlankLine
        \RETURN $\pinv{A}$
    \end{algorithmic}
\end{algorithm}

We propose \method (\methodlong), a novel method for efficiently and accurately computing the approximate pseudoinverse for sparse matrices. 
We describe the overall procedure of \method in Algorithm~\ref{alg:method}.
Our main ideas for accelerating the pseudoinverse computation are as follows:
\begin{itemize}
    \item {\textbf{Idea 1 (line~\ref{alg:method:reorder}).} Many feature matrices collected from real-world domains are highly sparse and skewed as shown in Figure~\ref{fig:degree_distributions} (\textbf{Section~\ref{sec:method:observation}}); and we show that these feature matrices can be reordered such that their non-zeros are concentrated as shown in Figure~\ref{fig:matrix_reordering:final} (\textbf{Section~\ref{sec:method:matrix_reordering}}).}
    \item {\textbf{Idea 2 (line~\ref{alg:method:svda11}).} The reordered matrix involves a large and sparse block diagonal submatrix whose SVD is easy-to-compute (\textbf{Section~\ref{sec:method:incremental_svd}}).
    }
    \item {\textbf{Idea 3  (lines~\ref{alg:method:row_inc_up}~and~\ref{alg:method:col_inc_up}).}
        The final SVD result of the feature matrix is efficiently obtained by incrementally updating the SVD result of the sparse submatrix (\textbf{Section~\ref{sec:method:incremental_svd}}).
    }
\end{itemize}

\subsection{Observation from Real-world Feature Matrix}
\label{sec:method:observation}
We first explain the skewness of feature matrices in the real-world datasets, which plays a key role in motivating the matrix reordering of \method.
A notable characteristic of feature matrices collected from many real-world problems is that they are extremely sparse as shown in Table~\ref{tab:dataset} (the details of the datasets are described in Section~\ref{sec:experiment}).

This sparsity naturally leads us to interpret $\mat{A}$ as a sparse network. 
Moreover, rows of a feature matrix map training instances, columns map the features, and non-zero values map the relations between instance-to-feature pairs.
Thus a feature matrix $\mat{A}$ naturally represents a bipartite network as follows:
\begin{definition}[Bipartite Network from Feature Matrix]
\label{definition:bipartite_network}
Given  $\mat{A} \in \R{m}{n}$, a bipartite network $G=(V_T, V_F, E)$ is derived from $\mat{A}$, where $V_T$ is the set of instance nodes, $V_F$ is the set of feature nodes, and $E$ is the set of edges between instance and feature nodes.
    For each non-zero entry $a_{ij}$, an edge $(i, j)$ is formed in $G$, where $i \in V_T$ and $j \in V_F$.
\end{definition}

\begin{figure}[!t]
    \centering
    \subfigure[The Amazon dataset]{
        \hspace{-4mm}
        \includegraphics[width=0.26\linewidth]{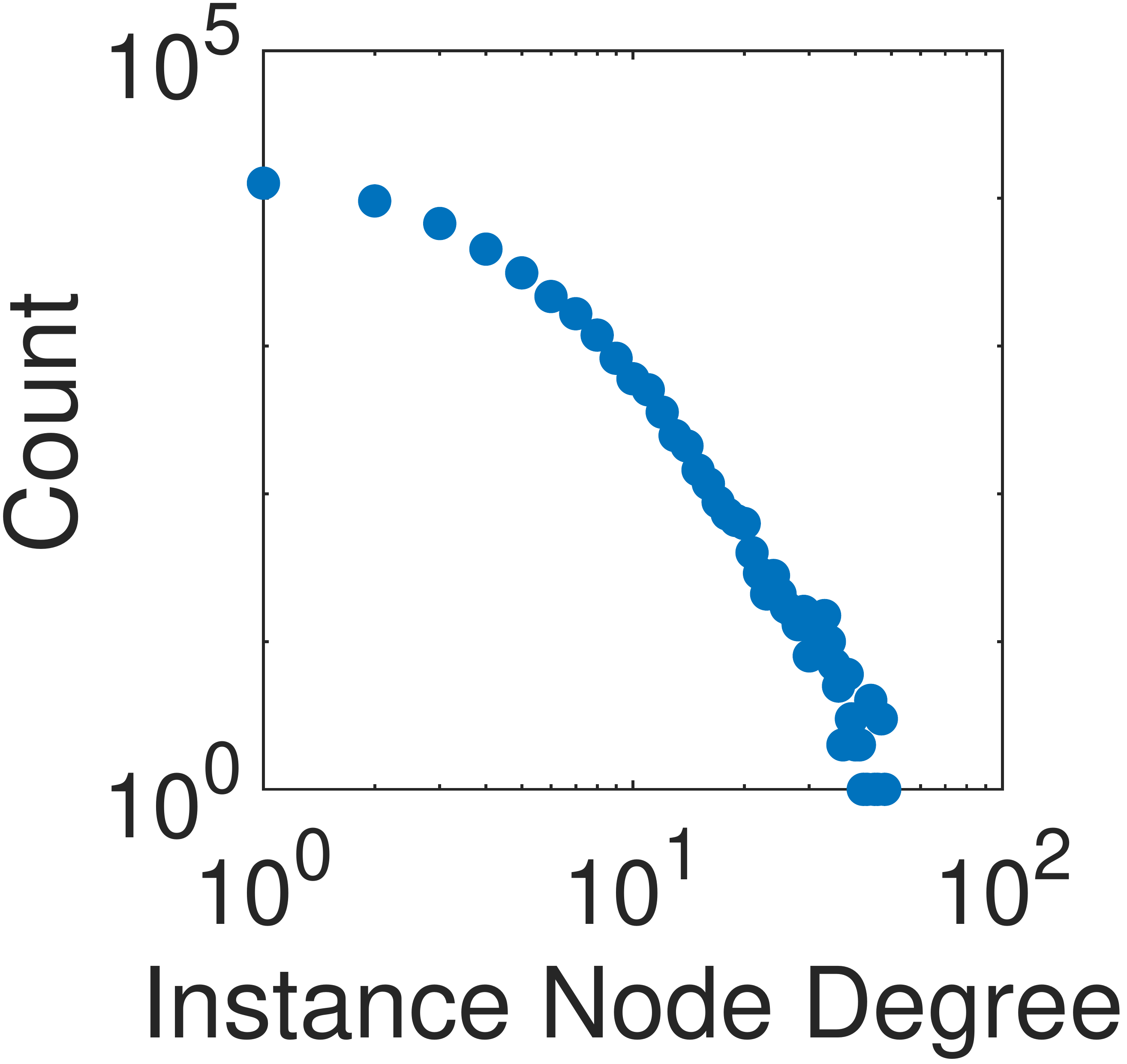}
        \hspace{-3mm}
        \includegraphics[width=0.255\linewidth]{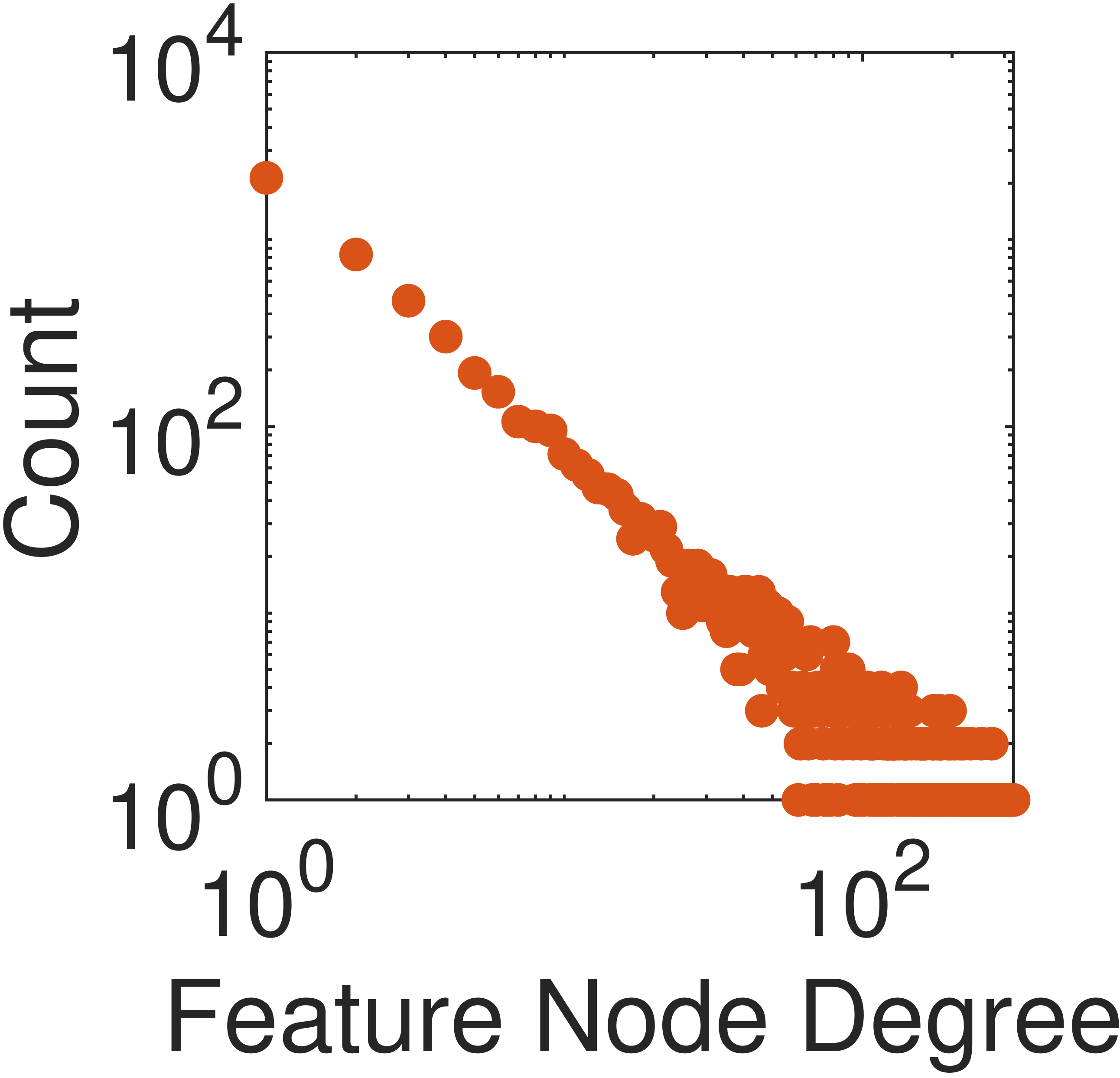}
    }
    \subfigure[The RCV dataset]{
        \hspace{-4mm}
        \includegraphics[width=0.26\linewidth]{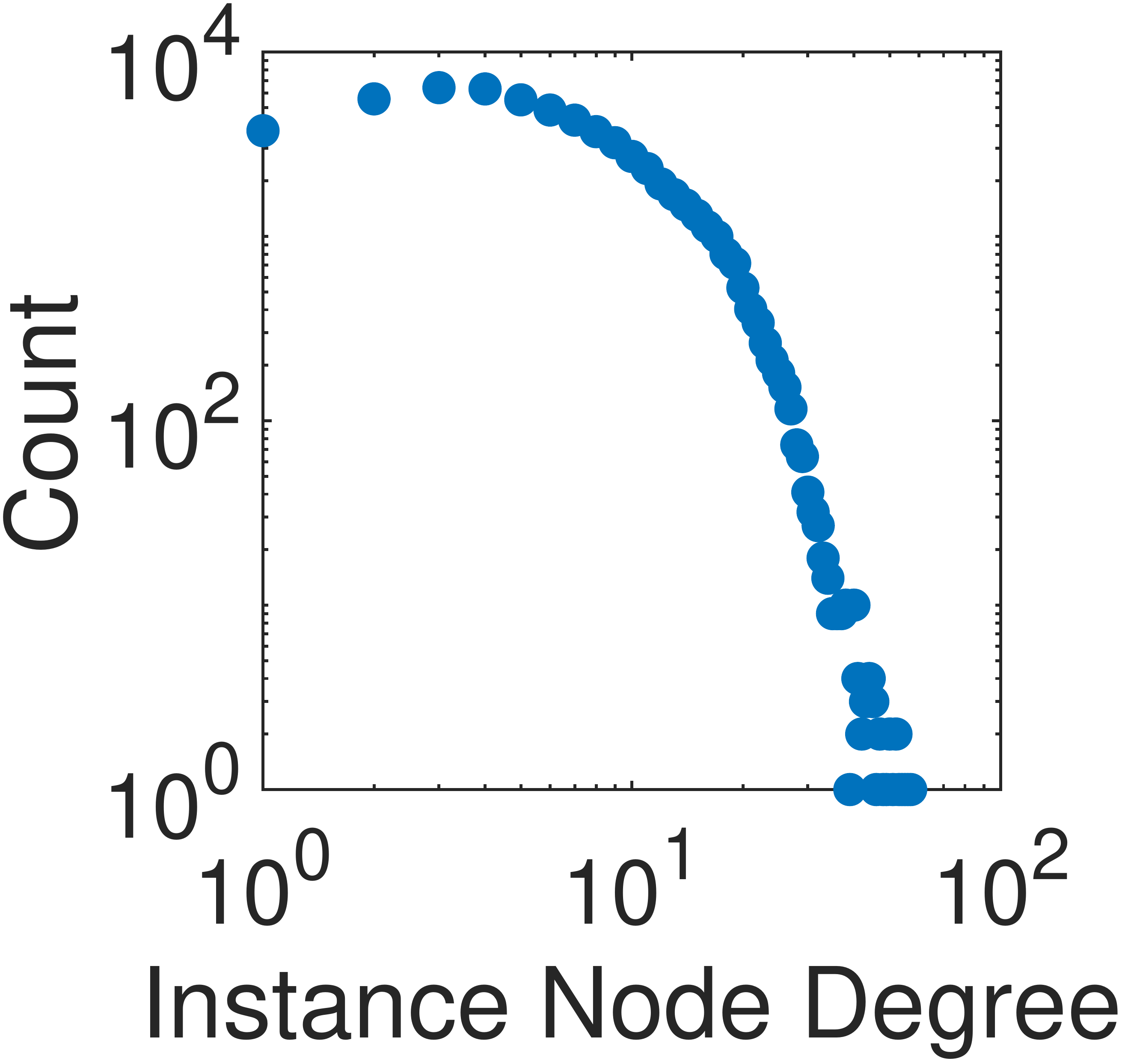}
        \hspace{-3mm}
        \includegraphics[width=0.255\linewidth]{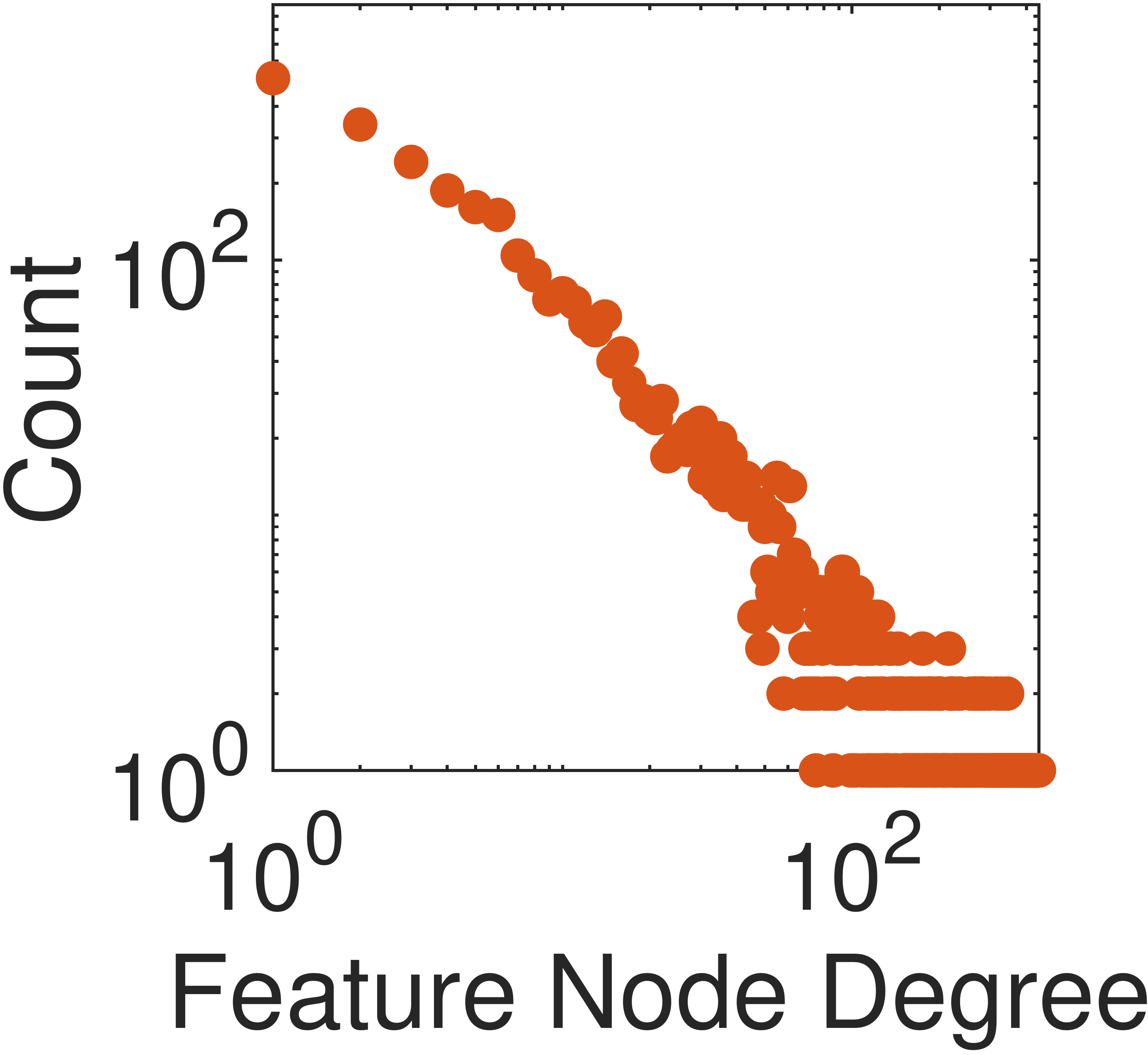}
        \hspace{-4mm}
    }
    \caption{
        \label{fig:degree_distributions}
        Degree distributions of instance and feature nodes in a bipartite network derived from real-world feature matrices.
        Note that there are few high degree nodes while the majority of nodes have low degrees, implying skewness on the degree distributions.
    }
\end{figure}

Figure~\ref{fig:degree_distributions} depicts the degree distributions of instance and feature nodes in each bipartite network derived from the Amazon and RCV feature matrices, respectively.
Note that the degree distributions of both bipartite networks
are skewed, i.e., there are only a few high degree nodes.
In network analysis, skewness of the degree distributions have been exploited to reorder association matrices for efficient analysis.  
In the network terminology, the high degree nodes are called \textit{hub} nodes, or simply hubs, and the neighbor nodes of a hub node are called the \textit{spoke} nodes, or simply spokes. There is no consistent threshold degree of a node for it to be considered a hub and often a relative proportion of high-degree nodes rather than an explicit threshold degree is used to select the hubs.

\begin{figure*}[!t]
    \centering
    \subfigure[Before removing hub nodes]{
        \label{fig:example_removing_hubs:before}
        \hspace{-1mm}
        \includegraphics[width=0.43\linewidth]{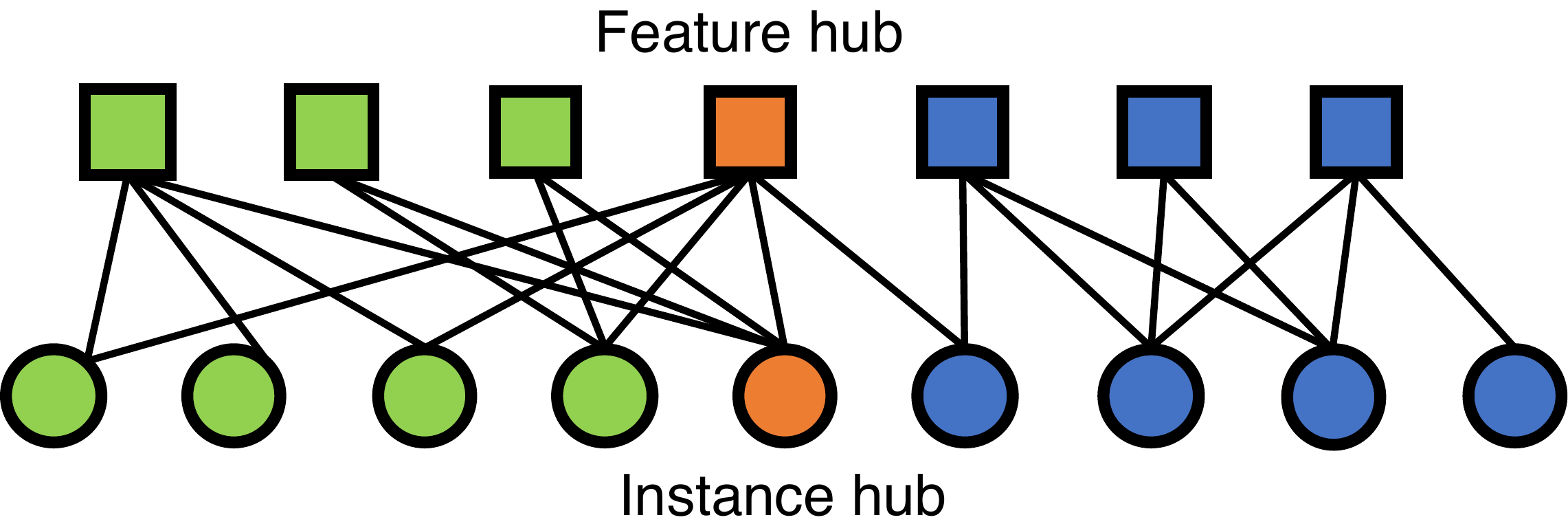}
        \hspace{2mm}
    }
    \subfigure[After removing hub nodes]{
        \label{fig:example_removing_hubs:after}
        \hspace{-1mm}
        \includegraphics[width=0.47\linewidth]{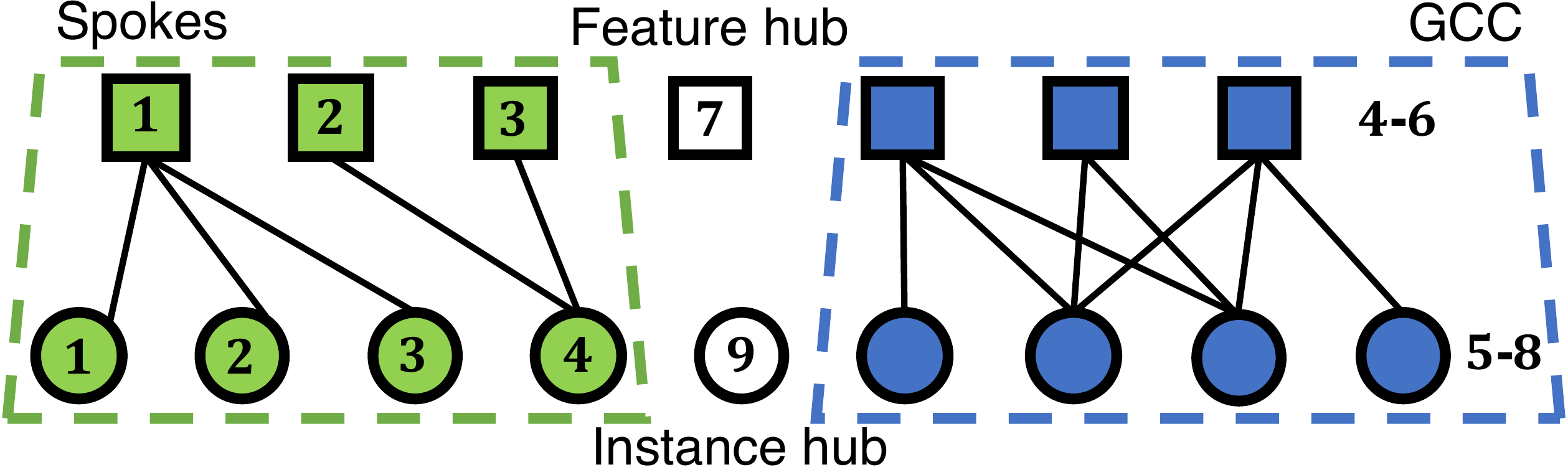}
        \hspace{-1mm}
    }
    \caption{
    \label{fig:example_removing_hubs}
    A bipartite network after one iteration of Algorithm~\ref{alg:matrix_reordering}, where a square indicates a feature node and a circle indicates an instance node.
    \method assigns the highest id to the feature node (id $7$) and instance node (id $9$), respectively.
    The nodes in spokes get the lowest ids and the GCC receives the remaining ids.
We remove multiple hub nodes in each iteration of the algorithm to sufficiently shatter the graph.
    }
\end{figure*}

Previous works on real-world networks have shown that real-world networks can be shattered by removing sets of highest degree nodes~\citep{KangF11,journals/tkde/LimKF14, Jung2016, Jung2017}. 
That is,  when a set of hubs is removed from a connected component, a non-trivial portion of the nodes, i.e., the \textit{spokes}, form small disconnected components, while the majority of the nodes remain in a giant connected component.
In figure~\ref{fig:example_removing_hubs}, we show that this shattering property of real-world networks also applies to bipartite graphs formed from feature matrices. We apply the shattering property to our feature matrix reordering (Algorithm~\ref{alg:matrix_reordering} of \method).



\subsection{Matrix Reordering of \method}
\label{sec:method:matrix_reordering}
\begin{figure*}[!t]
    \hspace{-3mm}
    \centering
    \subfigure[The original matrix]{
        \label{fig:matrix_reordering:original}
        \includegraphics[width=0.17\linewidth]{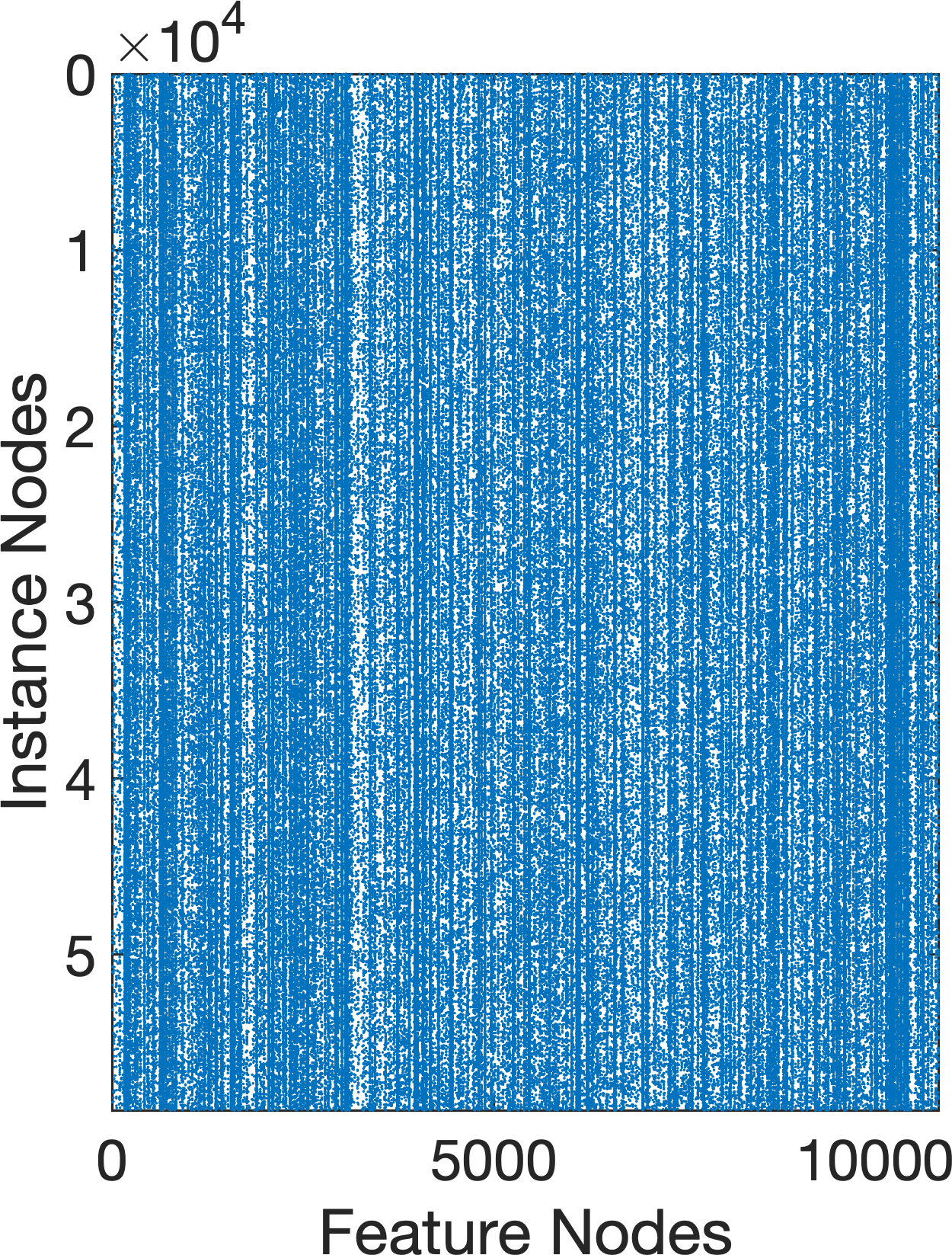}
    }
    \subfigure[After thirty iterations]{
        \label{fig:matrix_reordering:original_1}
        \includegraphics[width=0.17\linewidth]{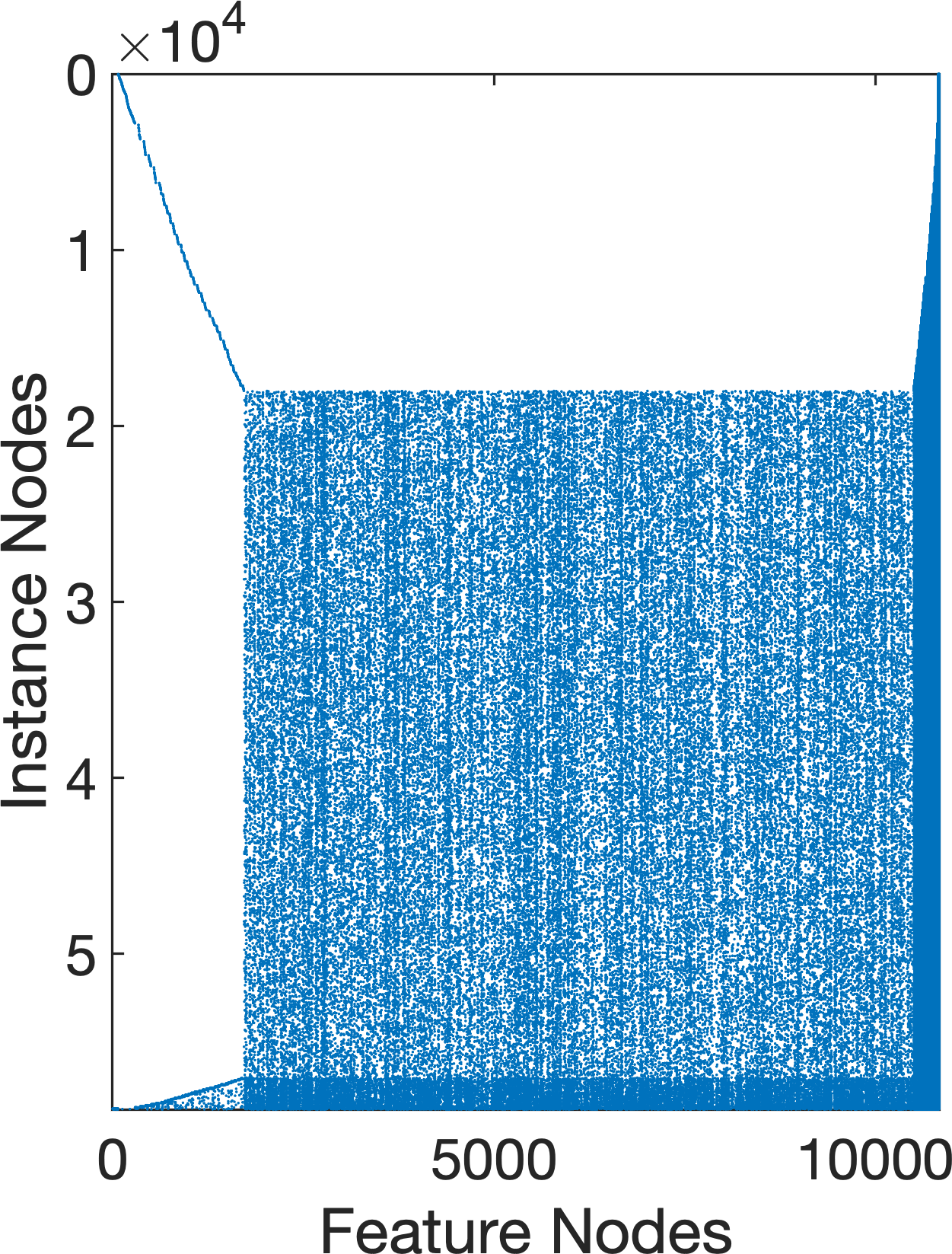}
    }
    \subfigure[After eighty iterations]{
        \label{fig:matrix_reordering:original_2}
        \includegraphics[width=0.17\linewidth]{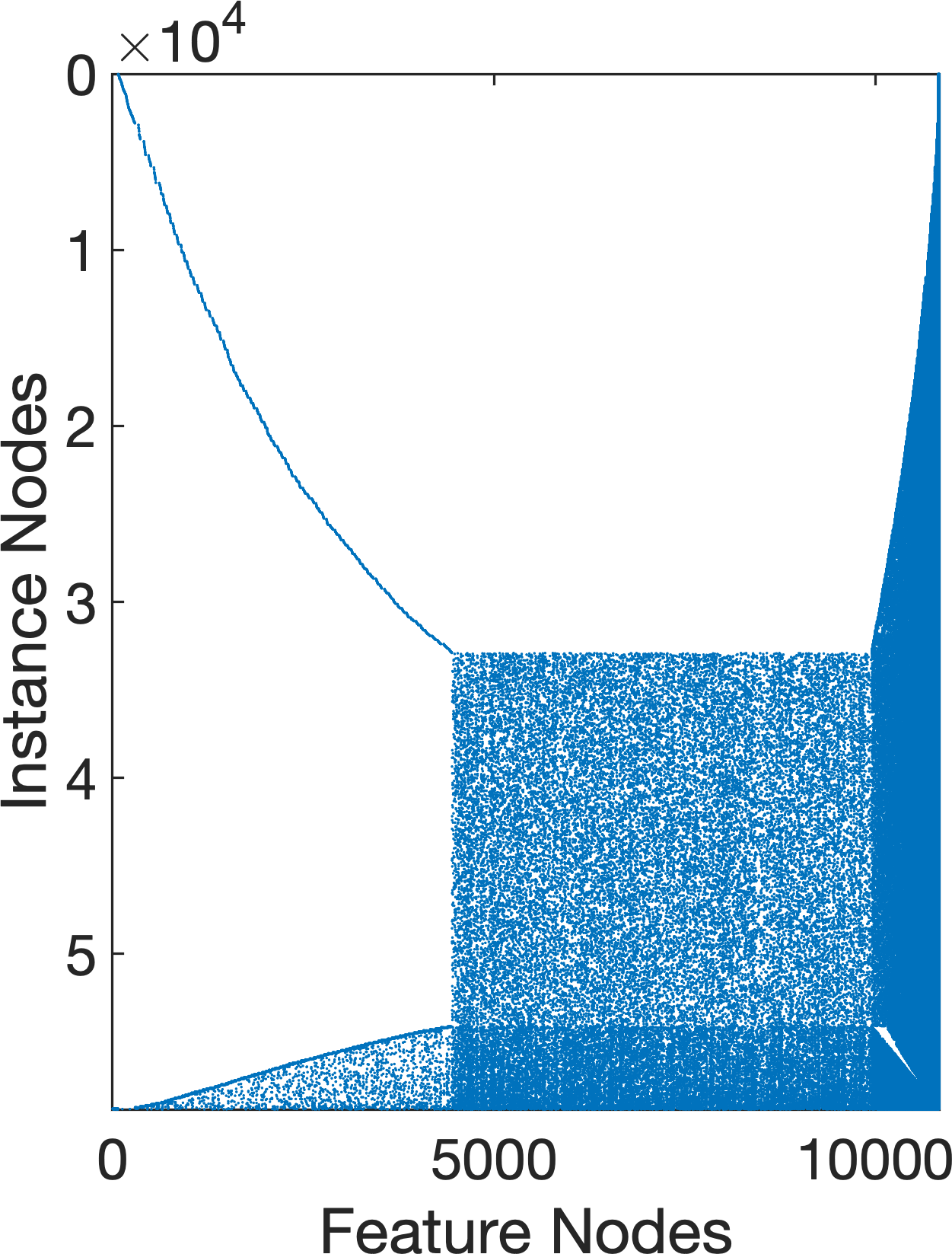}
    }
    \subfigure[After 115 iterations]{
        \label{fig:matrix_reordering:original_3}
        \includegraphics[width=0.17\linewidth]{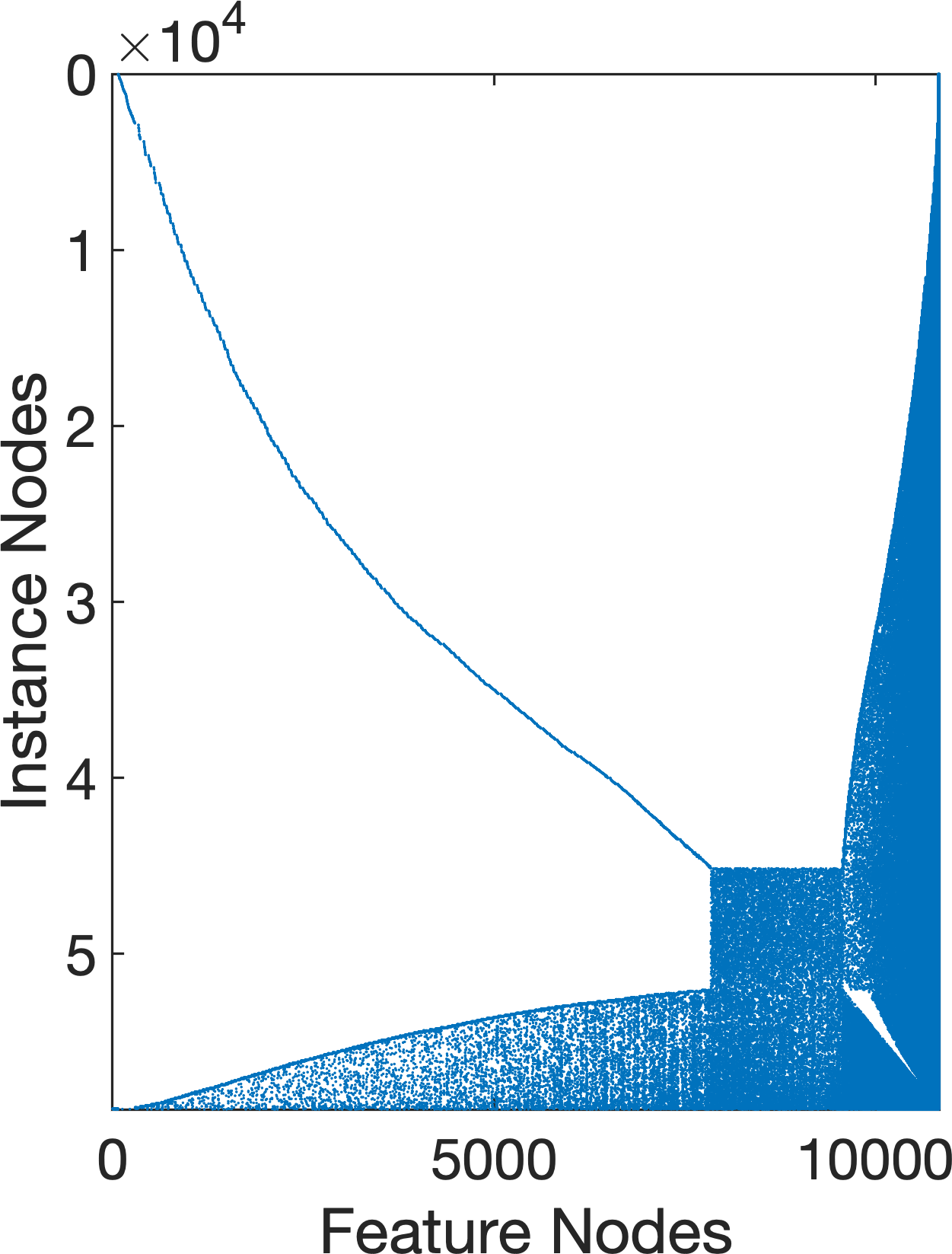}
    }
    \subfigure[After the final (119) iteration]{
        \label{fig:matrix_reordering:final}
        \includegraphics[width=0.19\linewidth]{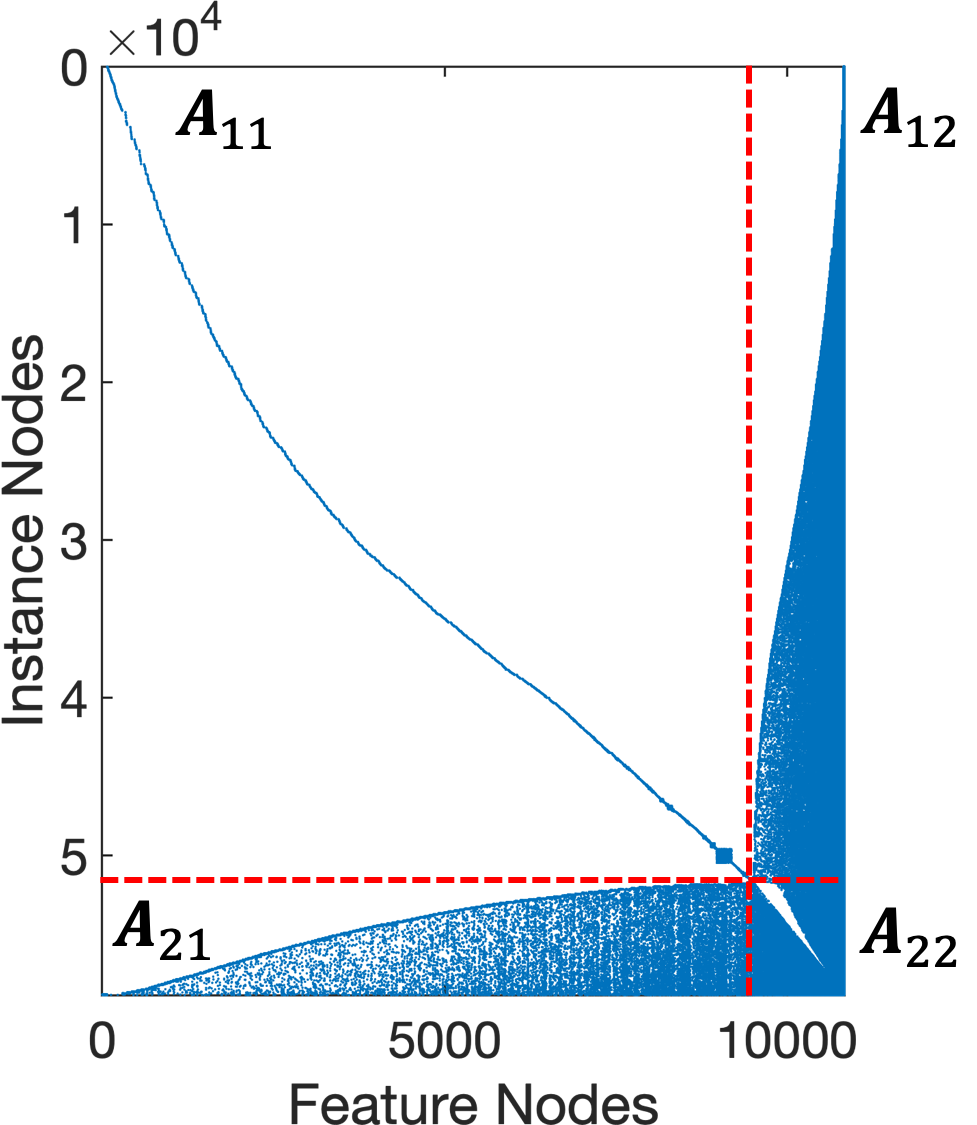}
    }
    \vspace{-3mm}
    \caption{
        \label{fig:matrix_reordering}
        The matrix reordering process of \method on the feature matrix of the Amazon dataset.
        (a) depicts the original matrix, (b-d) are the reordered matrix after several iterations in Algorithm~\ref{alg:matrix_reordering}, and (e) is the matrix after the final iteration.
        As shown in (e), the non-zero entries of the feature matrix are concentrated by the matrix reordering such that it is divided into four submatrices where $\An{11}$ is a large and sparse rectangular block diagonal matrix.
    }
    \vspace{-5mm}
\end{figure*}

Given a bipartite network $G$ derived from feature matrix $\mat{A}$ (Definition~\ref{definition:bipartite_network}), \method obtains permutation arrays $\pi_{T}: V_{T} \rightarrow \{1, \cdots, m\}$ for instance nodes and $\pi_{F} : V_{F} \rightarrow \{1, \cdots, n\}$ for feature nodes, such that the non-zero entries of the feature matrix are concentrated as seen in Figure~\ref{fig:matrix_reordering:final}.

The high-level mechanism of the matrix reordering procedure is summarized in Algorithm~\ref{alg:matrix_reordering}.
The algorithm first selects hub instance and feature nodes at line~\ref{alg:matrix_reordering:hub_selection} in the order of node degree; given a hub selection ratio $0 < k < 1$, it chooses $m_{\text{hub}} \leftarrow \lceil k\times|V_T| \rceil$ hub instance nodes and $n_{\text{hub}} \leftarrow \lceil k\times|V_F| \rceil$ hub feature nodes, respectively.
Then, it removes the selected hubs at line~\ref{alg:matrix_reordering:hub_removal}, such that the given network is split into three parts: 1) hubs, 2) giant connected component (GCC) (colored blue), and 3) spokes (colored green) to the hubs as shown in Figure~\ref{fig:example_removing_hubs}.

After removing the hubs, we assign new nodes ids to each $\pi_{T}$ and $\pi_{F}$ according to their node types.
In Figure~\ref{fig:example_removing_hubs},
the initial number of instance and features nodes are $|V_T|=9$ and $|V_F|=7$.
After line~\ref{alg:matrix_reordering:hub_removal},
the hub instance node gets the highest instance id 9 $(=|V_T|)$,
and the hub feature node gets the highest feature id 7 $(=|V_F|)$.
Note that those two hubs should be treated differently;
the instance node id corresponds to a row index, and the feature node id corresponds to a column index in the feature matrix.
At line~\ref{alg:matrix_reordering:find_spokes}, the nodes in spokes take the lowest ids as in Figure~\ref{fig:example_removing_hubs:after}.
The remaining ids are assigned to the GCC.
The same procedure is recursively repeated on the new GCC at line~\ref{alg:matrix_reordering:repeat_next_gcc}.

\begin{algorithm}[t!]
    \small
    \begin{algorithmic}[1]
        \caption{Matrix Reordering of \method}
        \label{alg:matrix_reordering}
        \REQUIRE bipartite network $G = (V_T, V_F, E)$ derived from feature matrix $\A$ and hub selection ratio $k$
        \ENSURE permutation arrays $\pi_{T}: V_{T} \rightarrow \{1, \cdots, m\}$ and $\pi_{F} : V_{F} \rightarrow \{1, \cdots, n\}$
        \REPEAT
        \STATE select $m_{\text{hub}} \leftarrow \lceil k\times|V_T| \rceil$ hubs in $V_T$ and $n_{\text{hub}} \leftarrow \lceil k\times|V_F| \rceil$ hubs in $V_F$, respectively \label{alg:matrix_reordering:hub_selection}
        \STATE place the selected hubs in $V_{T}$ and $V_{F}$ to the end of $\pi_{T}$ and $\pi_{F}$, respectively; and remove them from $G$ to generate new graph $G'$~ \label{alg:matrix_reordering:hub_removal}
        \STATE find the connected components in $G'$ using breadth first search; and place nodes belonging to each non-giant connected component at the beginning of $\pi_{T}$ and $\pi_{F}$, respectively \label{alg:matrix_reordering:find_spokes}
        \STATE set $G = (V_T, V_F, E)$ to be the giant connected component (GCC) of $G'$\label{alg:matrix_reordering:repeat_next_gcc}
        \UNTIL{the number of nodes in $V_{T}$ or $V_{F}$ of
        the GCC is smaller than current $m_{\text{hub}}$ or $n_{\text{hub}}$, respectively}
        \BlankLine
        \RETURN permutation arrays $\pi_{T}$ and $\pi_{F}$
    \end{algorithmic}
\end{algorithm}

Figure~\ref{fig:matrix_reordering} depicts the matrix reordering in Algorithm~\ref{alg:matrix_reordering} for the feature matrix of the Amazon dataset.
The spy plot of the feature matrix before reordering is in Figure~\ref{fig:matrix_reordering:original}.
Figures~\ref{fig:matrix_reordering:original_1}$\sim$\ref{fig:matrix_reordering:original_3} shows the intermediate matrices of the reordering process.
As iterations proceed, the non-zero entries of the feature matrix are concentrated at the bottom right corner of the feature matrix as shown in Figure~\ref{fig:matrix_reordering:final}.
The final reordered matrix can be divided into four submatrices, or blocks, where the top left submatrix is a large and sparse block diagonal matrix.
More specifically, the top and left submatrix contains small rectangular blocks at the diagonal area, where those blocks are formed by the spokes nodes; e.g., in Figure~\ref{fig:example_removing_hubs:after}, the feature node with id $1$ and instance nodes with ids $1$-$3$ are grouped to form a tiny rectangular block on the diagonal area of the submatrix.


\subsection{Incremental SVD Computation of \method}
\label{sec:method:incremental_svd}
The reordered matrix $\A$ is divided into \small$\mat{A} = \begin{bmatrix}
        \An{11} & \An{12} \\
        \An{21} & \An{22}
    \end{bmatrix}$, \normalsize\xspace
where $\An{11} \in \R{m_1}{n_1}$, $\An{12} \in \R{m_1}{n_2}$, $\An{21} \in \R{m_2}{n_1}$, and $\An{22} \in \R{m_2}{n_2}$.
Note that $m_1$ and $n_1$ are the number of spoke instance and feature nodes, respectively.
$m_2$ and $n_2$ are the number of hub instance and feature nodes, respectively.

\subsubsection{SVD Computation for $\An{11}$}

This step computes the low-rank SVD result of $\An{11}$.
Note that $\An{11}$ is large and sparse, where many but small rectangular blocks are located at the diagonal area of $\An{11}$ as shown in Figure~\ref{fig:matrix_reordering:final}.
In this case, SVD result of $\An{11}$ is efficiently obtained by computing SVD of each small block in $\An{11}$ instead of performing SVD on the whole submatrix.
For $i$th-block $\An{11}^{(i)} \in \R{m_{1i}}{n_{1i}}$, suppose $\mat{U}^{(i)}\mat{\Sigma}^{(i)}\mat{V}^{(i)\top}$ is the low-rank approximated SVD with the target rank $s_{i} = \lceil \alpha n_{1i} \rceil$ (let $m_{1i} > n_{1i}$ without loss of generality).
Then, the SVD result of $\An{11}$ is as follows:
\begin{align}
    \begin{split}
    \label{eq:block_diagonal}
    \U{m_1}{s}\S{s}{s}\VT{s}{n_1} =
    &\text{bdiag}(\mat{U}^{(1)}, \cdots, \mat{U}^{(B)}) \times \\
    & \text{bdiag}(\mat{\Sigma}^{(1)}, \cdots, \mat{\Sigma}^{(B)}) \times
    \text{bdiag}(\mat{V}^{(1)\top}, \cdots, \mat{V}^{(B)\top})
    \end{split}
\end{align}
\normalsize
where $B$ is the number of blocks and $\text{bdiag}(\cdot)$ is the function returning a rectangular block diagonal matrix with a valid block sequence.
The obtained rank is $s = \sum_{i=1}^{B}\alpha \lceil n_{1i} \rceil \approx \lceil \alpha n_1 \rceil$.
Note that this is also a valid SVD result since $\U{m_1}{s}$ and $\VT{s}{n_1}$ are orthogonal matrices, and $\S{s}{s}$ is diagonal, which follows the definition of SVD~\citep{strang2006linear}.

\subsubsection{Incremental Update of the SVD result}
The next step is to obtain the SVD result for $[\An{11}; \An{21}]$, where `$;$' indicates a vertical concatenation.
The SVD of $[\An{11}; \An{21}]$ is calculated by incremental SVD~\citep{brand2003fast,ross2008incremental} given the SVD of $\An{11}$.
The derivation for this incremental computation with the given target rank $s = \lceil \alpha n_1 \rceil$ is the followings:

\begin{align}
    \begin{split}
    \label{eq:row_incremental_update}
    \begin{bmatrix}
        \An{11} \\
        \An{21}
    \end{bmatrix}
    & \simeq
    \begin{bmatrix}
        \U{m_1}{s}\S{s}{s}\VT{s}{n_1} \\
        \An{21}
    \end{bmatrix}\\
    & =
    \begin{bmatrix}
        \U{m_1}{s} \!\! & \!\! \mat{O}_{m_1 \times m_2} \\
        \mat{O}_{m_2 \times s}  \!\! & \!\! \mat{I}_{m_2 \times m_2}
    \end{bmatrix}
    \begin{bmatrix}
        \S{s}{s}\VT{s}{n_1} \\
        \An{21}
    \end{bmatrix}
    \\
    &\simeq
    \begin{bmatrix}
        \U{m_1}{s} \!\! & \!\! \mat{O}_{m_1 \times m_2} \\
        \mat{O}_{m_2 \times s} \!\! & \!\! \mat{I}_{m_2 \times m_2}
    \end{bmatrix}
    \underbrace{
    \Ut{(s + m_2)}{s}\St{s}{s}\VtT{s}{n_1}
    }_\text{Low-rank approximation with $s$} \\
    &= \U{m}{s}\S{s}{s}\VT{s}{n_1}
    \end{split}
\end{align}
\normalsize
where  $\S{s}{s} = \St{s}{s}$,  $\VT{s}{n_1} = \VtT{s}{n_1}$, $\mat{O}$ is a zero matrix, and $\mat{I}$ is an identity matrix.
Note that
\small$\U{m}{s} = \begin{bmatrix}
        \U{m_1}{s} & \mat{O}_{m_1 \times m_2} \\
        \mat{O}_{m_2 \times s} & \mat{I}_{m_2 \times m_2}
    \end{bmatrix}
    \Ut{(s + m_2)}{s}$\normalsize\xspace is orthogonal since the product of two orthogonal matrices is also orthogonal~\citep{strang2006linear}.
Note also that any low-rank SVD algorithm can be used for this purpose;
we use frPCA~\citep{feng2018fast} for a given low target rank ($r < \lceil 0.3n \rceil$ used), and the standard SVD otherwise since frPCA is optimized for very low ranks, and thus it is too slow for handling high ranks.

The final step is to incrementally update the SVD result in equation~\eqref{eq:row_incremental_update} for
\small$\mat{T} = [\An{12} ; \An{22}]$\normalsize\xspace with $r = \lceil \alpha n \rceil$ as follows:
\begin{align}
    \begin{split}
    \label{eq:column_incremental_update}
        \begin{bmatrix}
            \An{11} \!\! & \!\! \An{12}\\
            \An{21} \!\! & \!\! \An{22}
        \end{bmatrix}
        & \simeq
        \begin{bmatrix}
            \U{m}{s}\S{s}{s}\VT{s}{n_1} \!\! & \!\! \mat{T}
        \end{bmatrix} \\
        &=
        \begin{bmatrix}
            \U{m}{s}\S{s}{s} & \mat{T}
        \end{bmatrix}
        \begin{bmatrix}
            \VT{s}{n_1} \!\! & \!\! \mat{O}_{s \times n_2} \\
            \mat{O}_{n_2 \times n_1}  \!\! & \!\! \mat{I}_{n_2 \times n_2}
        \end{bmatrix} \\
        &=
        \underbrace{
        \Ut{m}{r}\St{r}{r}\VtT{r}{(s+n_2)}
        }_{\text{Low-rank approximation with $r$}}
        \begin{bmatrix}
            \VT{s}{n_1} \!\! & \!\! \mat{O}_{s \times n_2} \\
            \mat{O}_{n_2 \times n_1} \!\! & \!\! \mat{I}_{n_2 \times n_2}
        \end{bmatrix} \\
        & =
        \U{m}{r}\S{r}{r}\VT{r}{n} 
    \end{split}
\end{align}
\normalsize
where $\S{r}{r} \! = \! \St{r}{r}$, $\U{m}{r} \! = \! \Ut{m}{r}$, and
\small$\VT{r}{n} = \VtT{r}{(s+n_2)}
        \begin{bmatrix}
            \VT{s}{n_1} \!\! & \!\! \mat{O}_{s \times n_2} \\
            \mat{O}_{n_2 \times n_1} \!\! & \!\! \mat{I}_{n_2 \times n_2}
        \end{bmatrix}$\normalsize\xspace are also orthogonal.


\subsection{Complexity Analysis}
We analyze the computational complexity of Algorithm~\ref{alg:method} in the following lemma:
\begin{lemma}\textnormal{\textbf{(Computational Complexity of \method)}}
\label{lemma:complexity}
    Given a feature matrix $\mat{A} \in \R{m}{n}$ and a target rank $r = \lceil \alpha n \rceil$, where $\alpha$ is the target rank ratio,
    the computational complexity of \method  is $O(mr^2 + n_1r^2 + m n_2 r + m_2 n_1 r + (\sum_{i=1}^{B} m_{1i}n_{1i}s_i) + T(m\log(m) + |\mat{A}|))$ prior to the final pseudoinverse construction (line~\ref{alg:method:final} in Algorithm~\ref{alg:method}),
    where $m_1$ and $n_1$ are the number of spoke instance and feature nodes, respectively,
    $m_2$ and $n_2$ are the number of hub instance and feature nodes, respectively,
    $B$ is the number of rectangular blocks,
    $m_{1i}$ and $n_{1i}$ are the height and the width of each rectangular block in $\An{11}$, respectively,
    $s_i = \lceil \alpha n_{1i} \rceil$ is the target rank of $i$-th block,
    $|\mat{A}|$ is the number of non-zeros of $\mat{A}$, and
$T$ is the number of iterations in Algorithm~\ref{alg:matrix_reordering}.
    \begin{proof}
		We summarize the complexity of each step of Algorithm~\ref{alg:method} in Table~\ref{tab:complexity}.
        For this proof, we use the traditional complexity of the low-rank approximation as describe in~\citep{gu1996efficient,halko2011finding}; for matrix $\mat{A} \in \R{p}{q}$, the low-rank approximation takes $O(pqk)$ time with target rank $k$.
        For a detailed comparison, we omit the cost of the final pseudoinverse construction (line 5 in Algorithm 1) because all SVD based methods should perform the construction as a common step.
        The complexity of each step of the algorithm is proved as follows:
            \begin{itemize}
                \item {Line~1: for each iteration, \method sorts the degrees of instance and feature nodes; thus, it requires up to $O(m\log(m))$ since $m > n$.
                    Then, it searches connected components in $G'$ using the breadth first search (BFS) algorithm in $O(|\A|)$ indicating the number of edges in the network.
                    Hence, each iteration demands $O(m\log(m) + |\A|)$ time.
                }
                \item {Line~2: \method computes the low-rank approximated SVD of each rectangular block in $O(m_{1i}n_{1i}s_i)$ with target rank $s_i = \lceil \alpha n_{1i} \rceil$; thus, it is $O(\sum_{i=1}^{B} m_{1i}n_{1i}s_i)$. }
                \item {Line~3: in equation~(2), the low-rank approximation takes $O((m_2 + s) n_1 s) = O(m_2n_1s + n_1s^2)$ time.
                    The matrix multiplication for $\U{m}{s}$ takes $O(m_1s^2)$ as follows:
                    \begin{equation*}
                        \U{m}{s} = \begin{bmatrix}
                        \U{m_1}{s} & \mat{O}_{m_1 \times m_2} \\
                        \mat{O}_{m_2 \times s} & \mat{I}_{m_2 \times m_2}
                        \end{bmatrix}
                        \Ut{(s + m_2)}{s} =
                        \begin{bmatrix}
                            \U{m_1}{s}\Ut{s}{s} \\
                            \Ut{m_2}{s}
                        \end{bmatrix}
                    \end{equation*}
                    \normalsize
                    where \small$\Ut{(s+m_2)}{s} = \begin{bmatrix}
                             \Ut{s}{s} \\
                            \Ut{m_2}{s}
                         \end{bmatrix}$\normalsize.
                    Since $s \leq r$, it is bounded by $O(m_1 r^2 + n_1 r^2 + m_2 n_1 r)$ where $s = \lceil \alpha n_1 \rceil$ and $r = \lceil \alpha n \rceil$ and $n_1 \leq n$.
                }
                \item{Line~4: in equation~(3), the low-rank approximation takes $O(m(n_2+s)r) = O(mn_2r + msr)$, and the matrix multiplication takes $O(n_1s^2)$ as follows:
                    \begin{align*}
                    	\begin{split}
                        \VT{r}{n} &= \VtT{r}{(s+n_2)}
                        \begin{bmatrix}
                            \VT{s}{n_1}  &  \mat{O}_{s \times n_2} \\
                            \mat{O}_{n_2 \times n_1}  &  \mat{I}_{n_2 \times n_2}
                        \end{bmatrix} =
                        \begin{bmatrix}
                            \VtT{r}{s}\VT{s}{n_1}  & \VtT{r}{n_2}
                        \end{bmatrix}
                        \end{split}
                    \end{align*}
                    
                    where  \small$\VtT{r}{(s+n_2)} \! = \! \begin{bmatrix}
                             \VtT{r}{s} \!\! & \!\! \VtT{r}{n_2}
                         \end{bmatrix}$\normalsize.
                    Hence, it is $O(n_1r^2 + mr^2 + mn_2r)$ since $s \leq r$.\QEDB
               }
    	\end{itemize}
    \end{proof}
\end{lemma}

\def\arraystretch{1.1}
\setlength{\tabcolsep}{6pt}
\begin{table}[!t]
\vspace{3mm}
\begin{threeparttable}[t]
    \caption{
        Computational complexity of each step of \method (Algorithm~\ref{alg:method}).
    }
\begin{tabular}{c | l | l}
\toprule
\multicolumn{1}{c|}{\textbf{Line}} & \multicolumn{1}{c|}{\textbf{Task}}   & \multicolumn{1}{c}{\textbf{Computational Complexity}} \\
\midrule
1                     & Reorder $\A$ using Algorithm~2                  & $O(T(m\log(m) + |\A|))$                            \\
2                     & Compute SVD of $\An{11}$         & $O(\sum_{i=1}^{B} m_{1i}n_{1i}s_i)$                                     \\
3                     & Update the SVD result for $\An{21}$ & $O(m_1 r^2 + n_1 r^2 + m_2 n_1 r)$                          \\
4                     & Update the SVD result for $\mat{T}=[\An{12}; \An{22}]$   & $O(n_1r^2 + mr^2 + mn_2r)$                                        \\
\midrule
\textbf{Total}                    &  \multicolumn{2}{l}{$O(mr^2 + n_1r^2 + m n_2 r + m_2 n_1 r + (\sum_{i=1}^{B} m_{1i}n_{1i}s_i) + T(m\log(m) + |\mat{A}|))$}                                           \\
\bottomrule
\end{tabular}
\label{tab:complexity}
\end{threeparttable}
\end{table}

The dominant factor is $mr^2$ in the complexity of the analysis of \method .
As described in Section~\ref{sec:preliminaries}, $O(mr^2)$ is faster than $O(mnr)$ of traditional methods~\citep{gu1996efficient}.
\method exhibits similar complexity to Randomized SVD, the state-of-the-art method with complexity of $O(mr^2 + nr^2 + mn\log(r))$~\citep{halko2011finding}.
However, the actual running time of the Randomized SVD is slower than that of the \method for a reasonably high rank (see Figure~\ref{fig:running_time}).
The reason is that Randomized SVD is based on oversampling technique (see the details in Section~\ref{sec:experiment:setting}); due to this point, Randomized SVD has a higher coefficient for the dominant factor compared to \method (i.e., Randomized SVD requires $4mr^2$ operations while \method needs $mr^2$ ones for the same $r$).

Note that for a detailed comparison, we have omitted the cost of the final pseudoinverse construction (line~\ref{alg:method:final} in Algorithm~\ref{alg:method}), which is a universal step in all SVD based methods.


%% file: 004experiment.tex
\def\arraystretch{1.2}
\setlength{\tabcolsep}{4.7pt}
\begin{table*}[!]
\small
\vspace{3mm}
\begin{threeparttable}[t]
    \caption{
        Dataset statistics.
        $m$ is the number of instances (rows), and
        $n$ is the number of features (columns) of feature matrix $\mat{A}$.
        $L$ is the number of labels of label matrix $\mat{Y}$.
        $|\mat{A}|$ is the number of non-zero entries of $\mat{A}$, and
        $\text{sp}(\mat{A})$ is the sparsity of $\mat{A}$ defined in Section~\ref{sec:experiment:setting}.
        $k$ is the hub selection ratio of Algorithm~\ref{alg:matrix_reordering}.
        $m_2$ and $n_2$ are the number of instance and feature hub nodes, respectively.
    }
\begin{tabular}{l | rrrrrrrrr}
\toprule
\textbf{Dataset}   & $m$     & $n$     & $L$     & $|\mat{A}|$  & $\text{sp}(\mat{A})$ & $\text{sp}(\mat{Y})$ & $k$    & $m_2$   & $n_2$   \\
\midrule
\textbf{Amazon} & 59,312 & 10,195 & 13,330 & 167,015  & 0.9997   &  0.9996   & 0.01 & 4,158 & 714 \\
\textbf{RCV}       & 62,385 & 4,724  & 2,456  & 466,675  & 0.9984    & 0.9981   & 0.01 & 8,112 & 624   \\
\textbf{Eurlex}    & 15,539 & 5,000  & 3,993  & 3,684,773 & 0.9525  & 0.9987   & 0.01 & 8,736 & 2,800   \\
\textbf{Bibtex}    & 7,395  & 1,836  & 159   & 507,746  & 0.9626   & 0.9849    & 0.01 & 5,180  & 1,330   \\
\bottomrule
\end{tabular}
\label{tab:dataset}
\end{threeparttable}
\end{table*}

In this section, we aim to answer the following questions from experiments:
\begin{itemize}
	\item {
		\textbf{Q1. Reconstruction error (Section~\ref{sec:experiment:recon_error}).}
		Does \method correctly produce low-rank SVD results in terms of reconstruction error?
	}
	\item {\textbf{Q2. Accuracy (Section~\ref{sec:experiment:accuracy}).}
        How accurate is the pseudoinverse obtained by \method for the multi-label linear regression task compared to other methods?
    }
    \item {\textbf{Q3. Efficiency (Section~\ref{sec:experiment:computational}).}
    	How quickly does \method compute the approximate pseudoinverse of sparse feature matrices compared to state-of-the-art methods?}
\end{itemize}


\subsection{Experimental Setting}
\label{sec:experiment:setting}

\textbf{Datasets.}
We use four real-world multi-label datasets, and their statistics are summarized in Table~\ref{tab:dataset}.
The Bibtex dataset is from a social bookmarking system, where each instance consists of features from a bibtex item and labels are tags in the system~\citep{katakis2008multilabel}.
The Eurlex dataset is from doc­u­ments about Eu­ro­pean Union law, where each instance is formed by word features from a document and labels indicate categories~\citep{mencia2008efficient}.
The RCV dataset is randomly sampled from an archive of newswire stories made available by Reuters, Ltd., where each instance consists of features from a document, and labels are categories~\citep{lewis2004rcv1}.
The Amazon dataset is randomly sampled from a set of reviews of Amazon, where each instance is formed by word features of a review and labels are items~\citep{mcauley2013hidden}. 
In Table~\ref{tab:dataset}, $\text{sparsity}(\mat{A})$ indicates the sparsity of feature matrix $\mat{A} \in \R{m}{n}$, which is defined as $\text{sp}(\mat{A}) = 1 - |\A|/(mn)$, where $|\mat{A}|$ is the number of non-zero entries in $\mat{A}$.

\textbf{Machine and Implementation.}
We use a single thread in a machine with an Intel Xeon E5-2630 v4 2.2GHz CPU and 512GB RAM.
All tested methods, including our proposed \method, are implemented using MATLAB which provides a state-of-the-art linear algebra package.

\textbf{Competing Methods.}
We use the following methods as the competitors of \method:
\begin{itemize}
\item{
\textbf{RandPI} is based on Randomized SVD~\citep{halko2011finding}, the state-of-the-art low-rank SVD method for target rank $r=\lceil \alpha n \rceil$ using an oversampling technique for numerical stability.
The main procedure of RandPI is as follows:
\begin{itemize}
	\item Step 1: It generates a Gaussian over-sampled random matrix $\mat{X}_{n \times 2r}$ for constructing randomized data matrix $\mat{B}_{m \times 2r} = \mat{A}\mat{X}_{n \times 2r}$.
	\item Step 2: It finds a matrix $\mat{Q}_{m \times 2r}$ with orthonormal columns as a proxy of an orthogonal matrix from $\mat{B}_{m \times 2r}$ satisfying $\mat{A} \simeq \mat{Q}\matt{Q}\mat{A}$.
	\item Step 3: It constructs $\mat{Y}_{2r \times n} = \matt{Q}_{2r \times m}\mat{A}$ and compute SVD of $\mat{Y}_{2r \times n} \simeq \Ut{2r}{r}\S{r}{r}\VT{r}{n}$.
	\item Step 4: It computes $\mat{U}_{m \times r} =\mat{Q}_{m \times 2r}\Ut{2r}{r}$.
\end{itemize}
\item{
\textbf{KrylovPI} is based on a Krylov subspace iterative method for the low-rank SVD method~\citep{baglama2005augmented}.
KrylovPI is specialized for computing a few singular values and vectors on a sparse matrix (used in \texttt{svds} of MATLAB).
}
\item{
\textbf{frPCA}~\citep{feng2018fast} combines the randomized SVD and a power iteration method so that it controls trade-off between running time and accuracy for computing SVD of sparse data. 
This also exploits LU decomposition in the power iteration to improve accuracy. 
We set the over-sampling parameter $s$ to $5$ and the number of iterations to $11$ as in~\citep{feng2018fast}.
}
}
\end{itemize}

\subsection{Reconstruction Error}
\label{sec:experiment:recon_error}
We analyze the reconstruction error of the SVD result computed by each method to check if it computes the SVD result accurately.
The reconstruction error of the SVD result $\mathbf{U}_{m \times r}\mathbf{\Sigma}_{r \times r}\mathbf{V}^{\top}_{r \times n}$ from the original matrix $\mat{A}$ is defined as  $\lVert \mathbf{A} - \mathbf{U}_{m \times r}\mathbf{\Sigma}_{r \times r}\mathbf{V}^{\top}_{r \times n}\rVert_{\text{F}}$, where $r=\lceil \alpha n \rceil$ is the target rank and $\lVert \cdot \rVert_{\text{F}}$ is the Frobenius norm.
We measure the error of each method varying the target rank ratio $\alpha$ from $0.01$ to $1.0$, respectively.

\begin{figure*}[!t]
    \centering
    \subfigure[Amazon]{
        \hspace{-4mm}
        \includegraphics[width=0.243\linewidth]{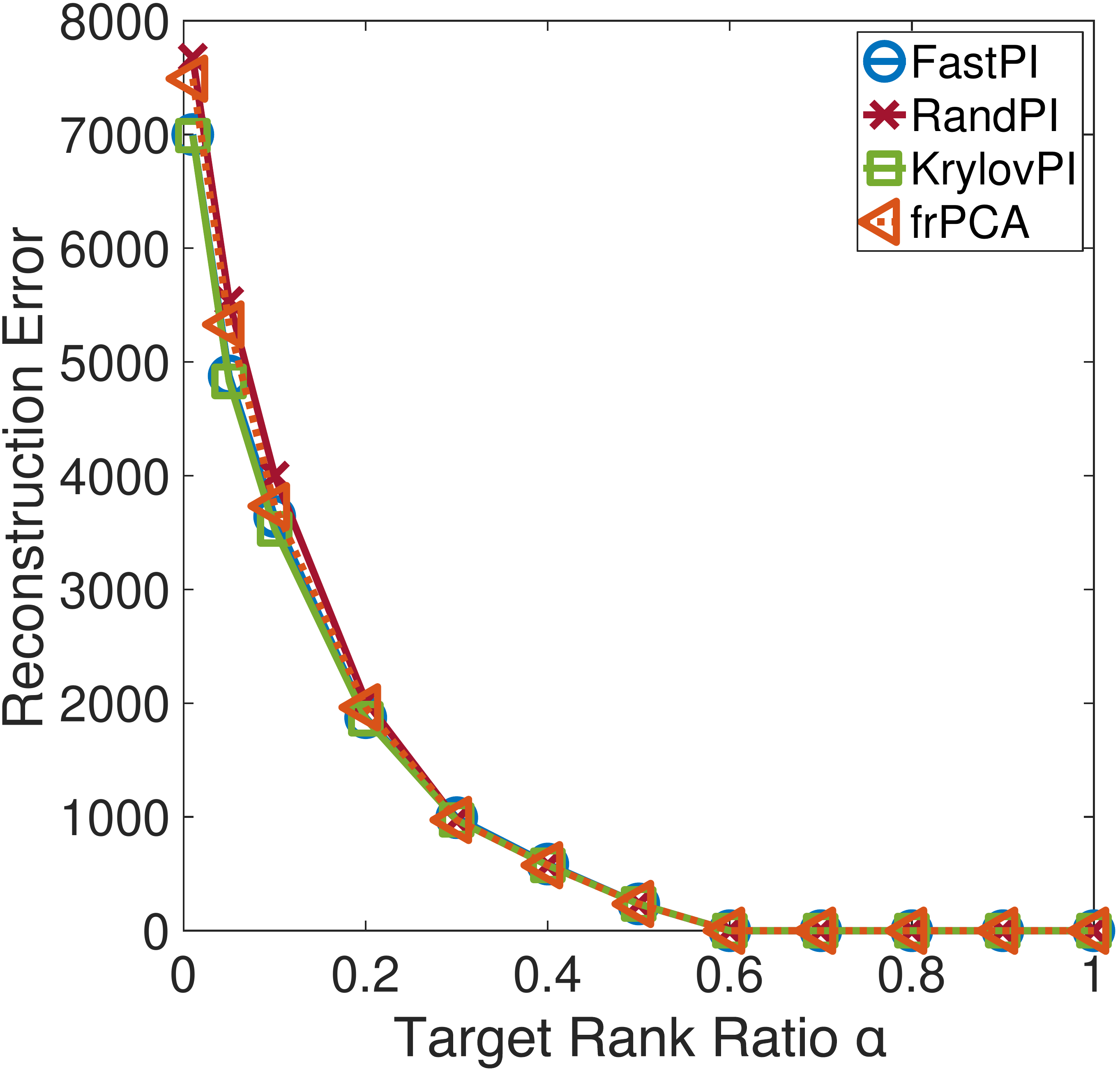}
    }
    \subfigure[RCV]{
        \hspace{-2mm}
        \includegraphics[width=0.232\linewidth]{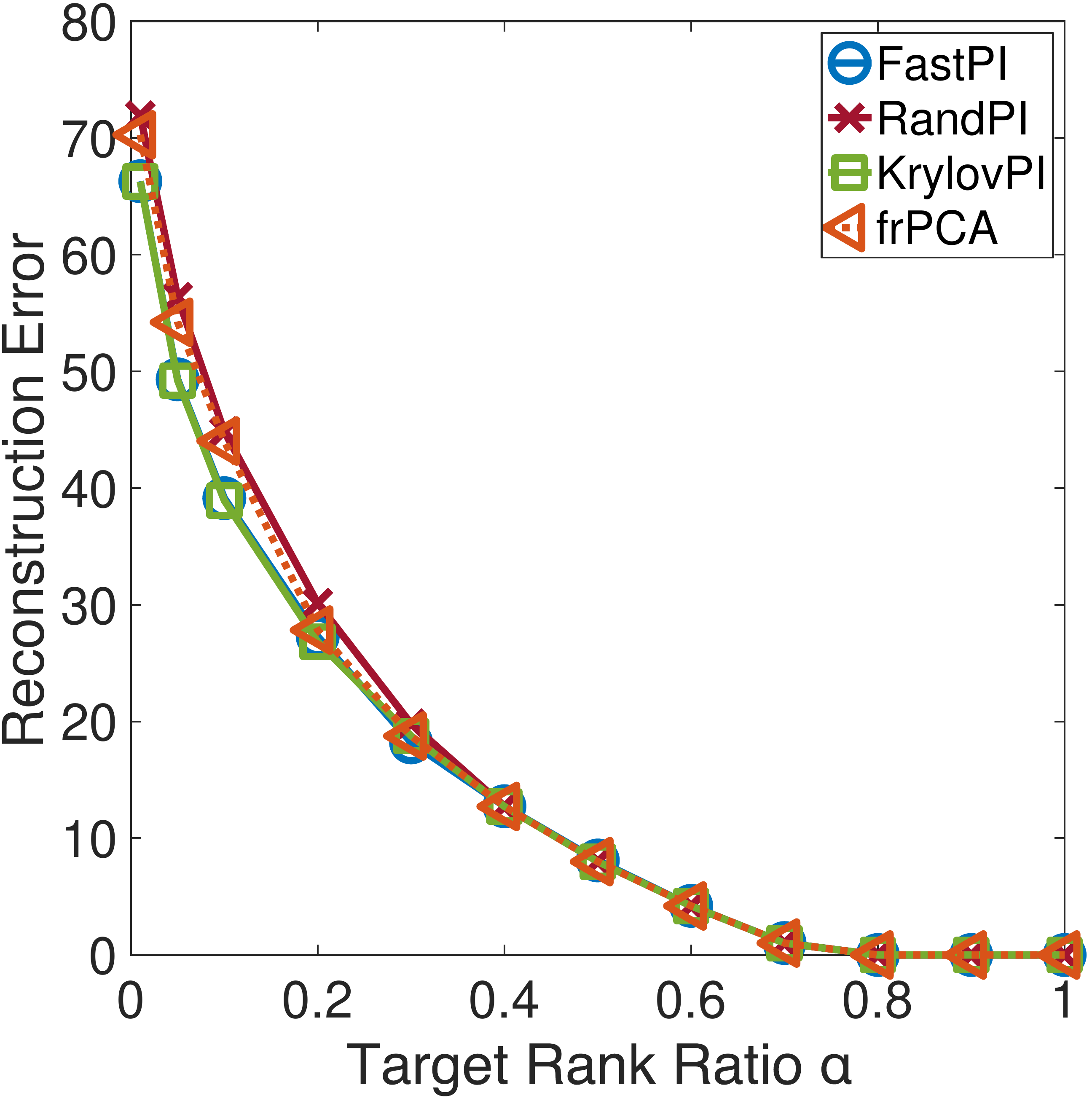}
      }
      \subfigure[Eurlex]{
        \hspace{-2mm}
        \includegraphics[width=0.223\linewidth]{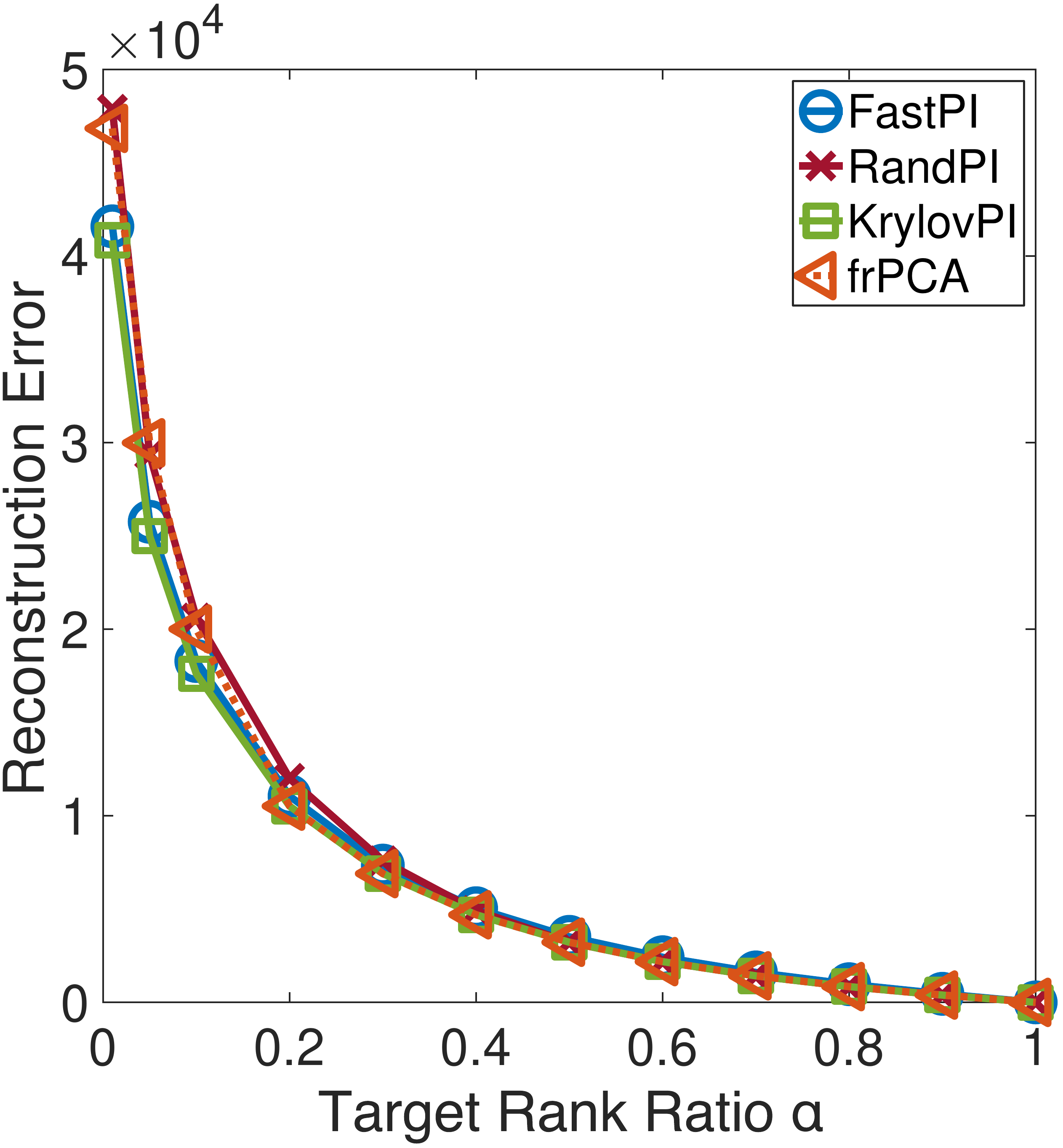}
      }
      \subfigure[Bibtex]{
        \hspace{-2mm}
        \includegraphics[width=0.237\linewidth]{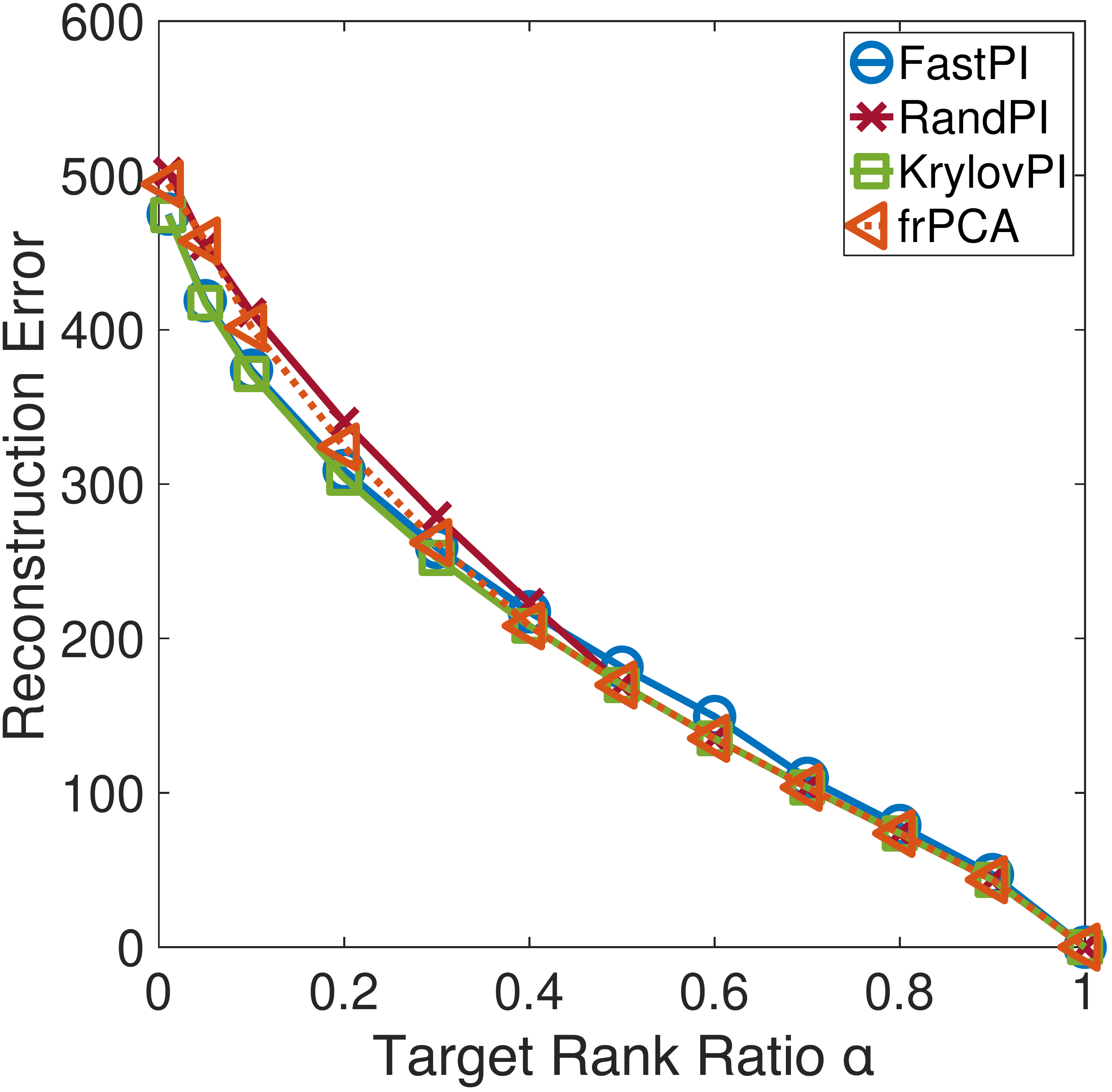}
      }
    \caption{
        \label{fig:error}
        Reconstruction error of the SVD result of each method varying target rank ratio $\alpha$ (Section~\ref{sec:experiment:recon_error}).
        Note that the error of our \method is almost the same as that of KrylovPI, indicating that our method computes the SVD result near optimally for any $\alpha$.
    }
    \vspace{-3mm}
\end{figure*}

Figure~\ref{fig:error} demonstrates the reconstruction error of all tested methods. 
The error of our FastPI is slightly better than that of RandPI, which is the state-of-the-art SVD for low-rank SVD computation, especially when $\alpha$ is low.
Another point is that the reconstruction error of our method is almost the same as that of KrylovPI.
These show that reconstruction-wise, our method is near-optimal given rank ratio $\alpha$ for the SVD computation.

\begin{figure*}[!t]
    \centering
    \subfigure[Amazon]{
        \hspace{-4mm}
        \includegraphics[width=0.232\linewidth]{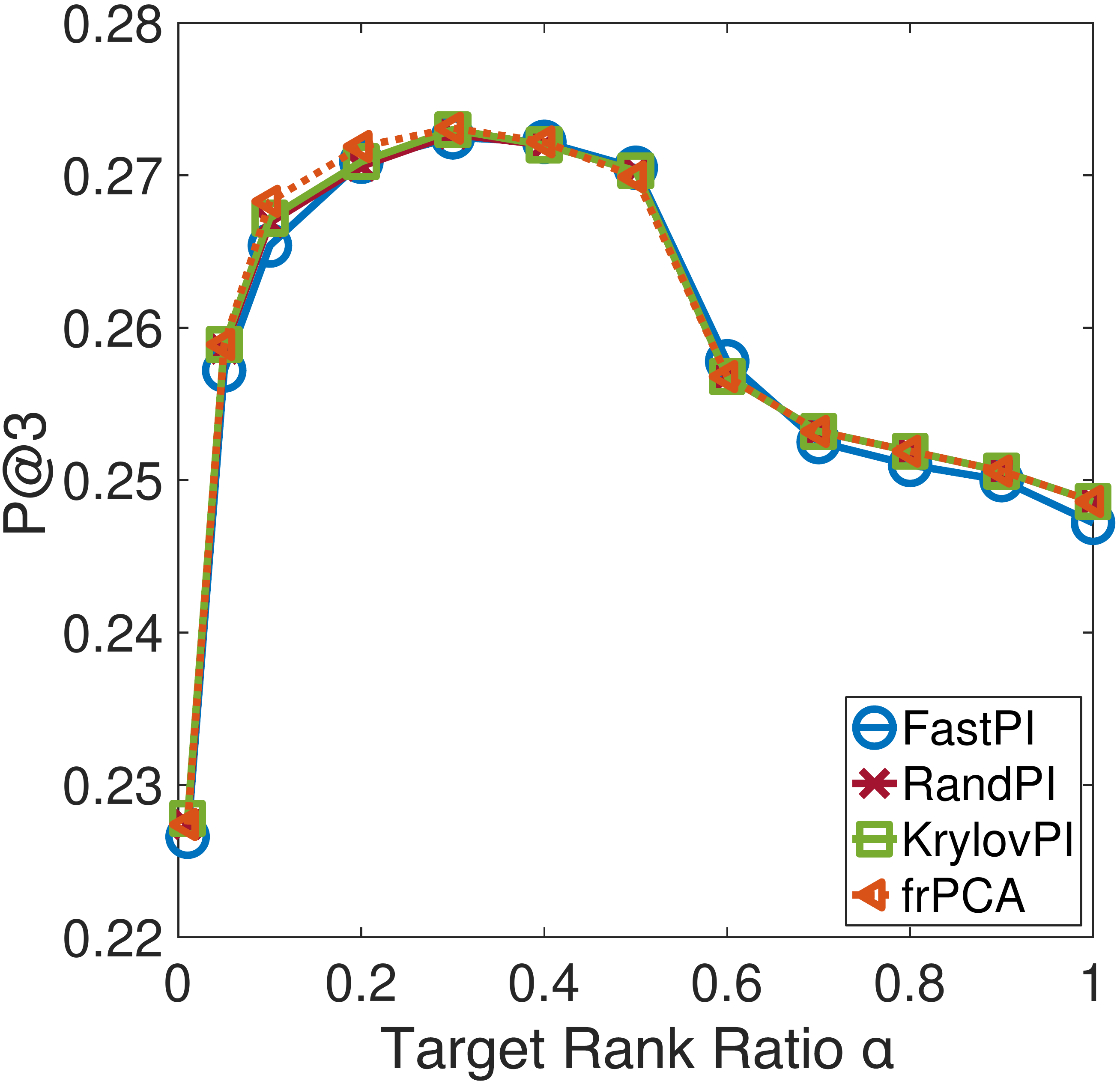}
    }
    \subfigure[RCV]{
        \hspace{-2mm}
        \includegraphics[width=0.232\linewidth]{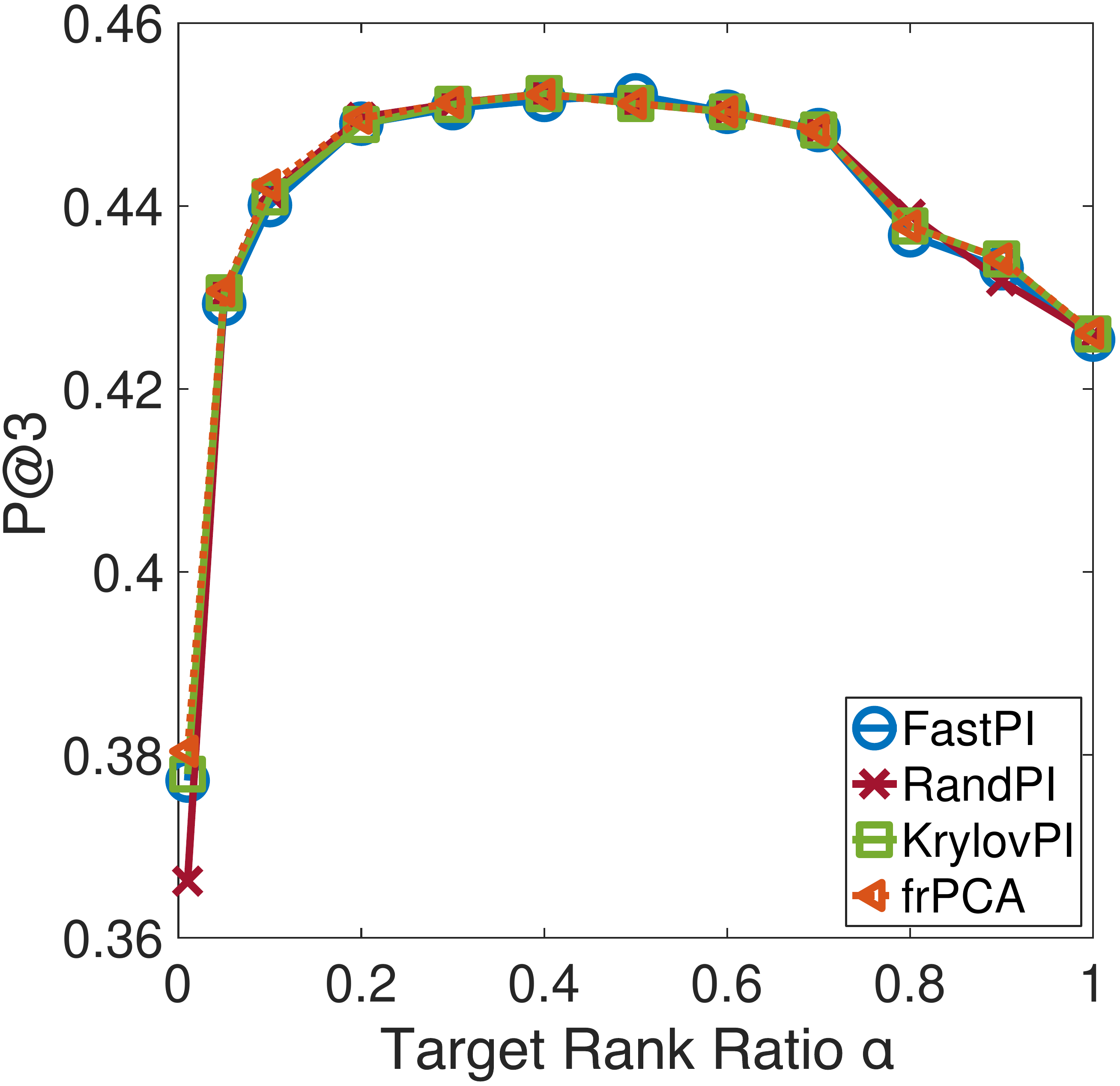}
      }
      \subfigure[Eurlex]{
        \hspace{-2mm}
        \includegraphics[width=0.225\linewidth]{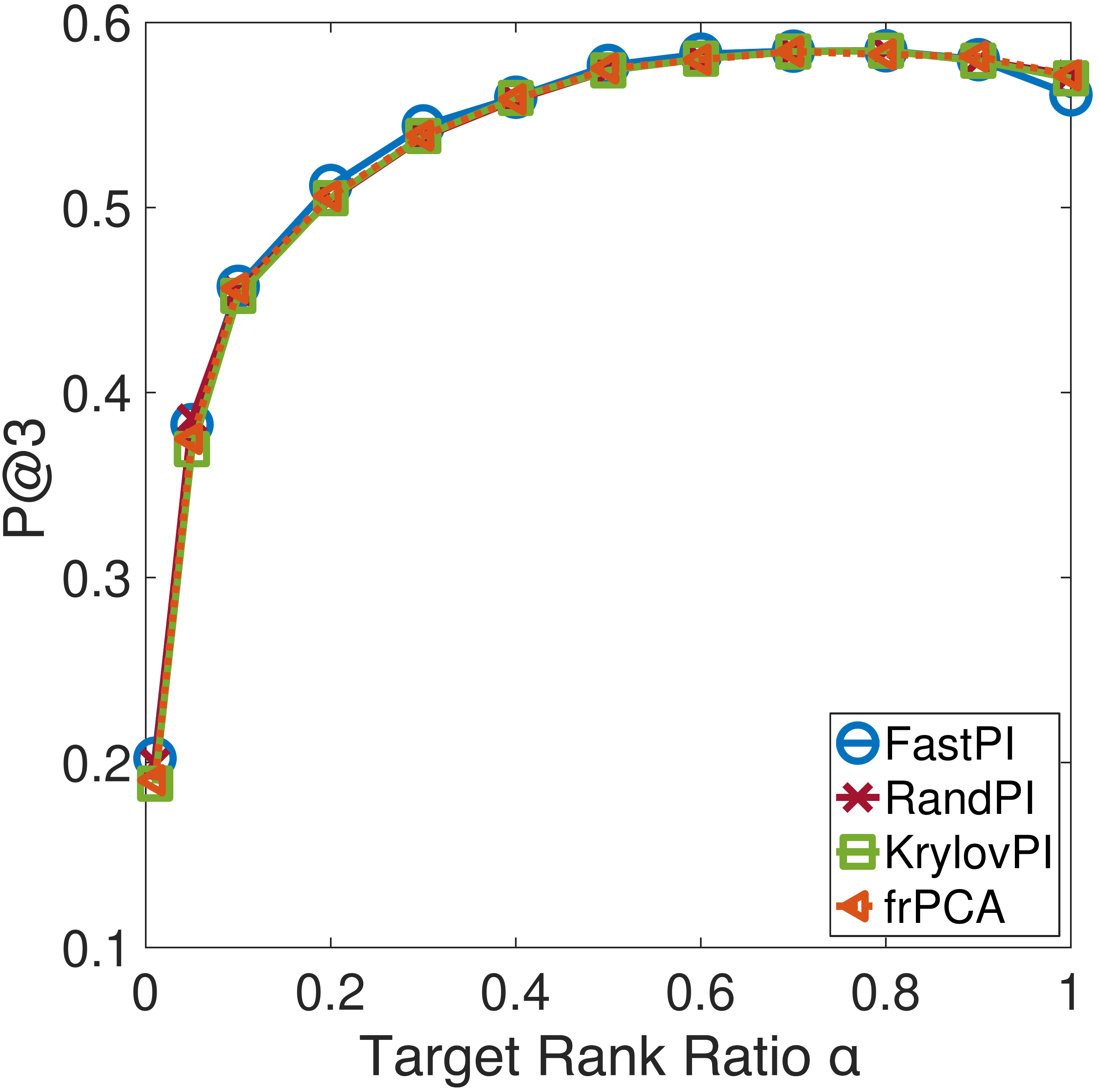}
    }
    \subfigure[Bibtex]{
        \hspace{-2mm}
        \includegraphics[width=0.23\linewidth]{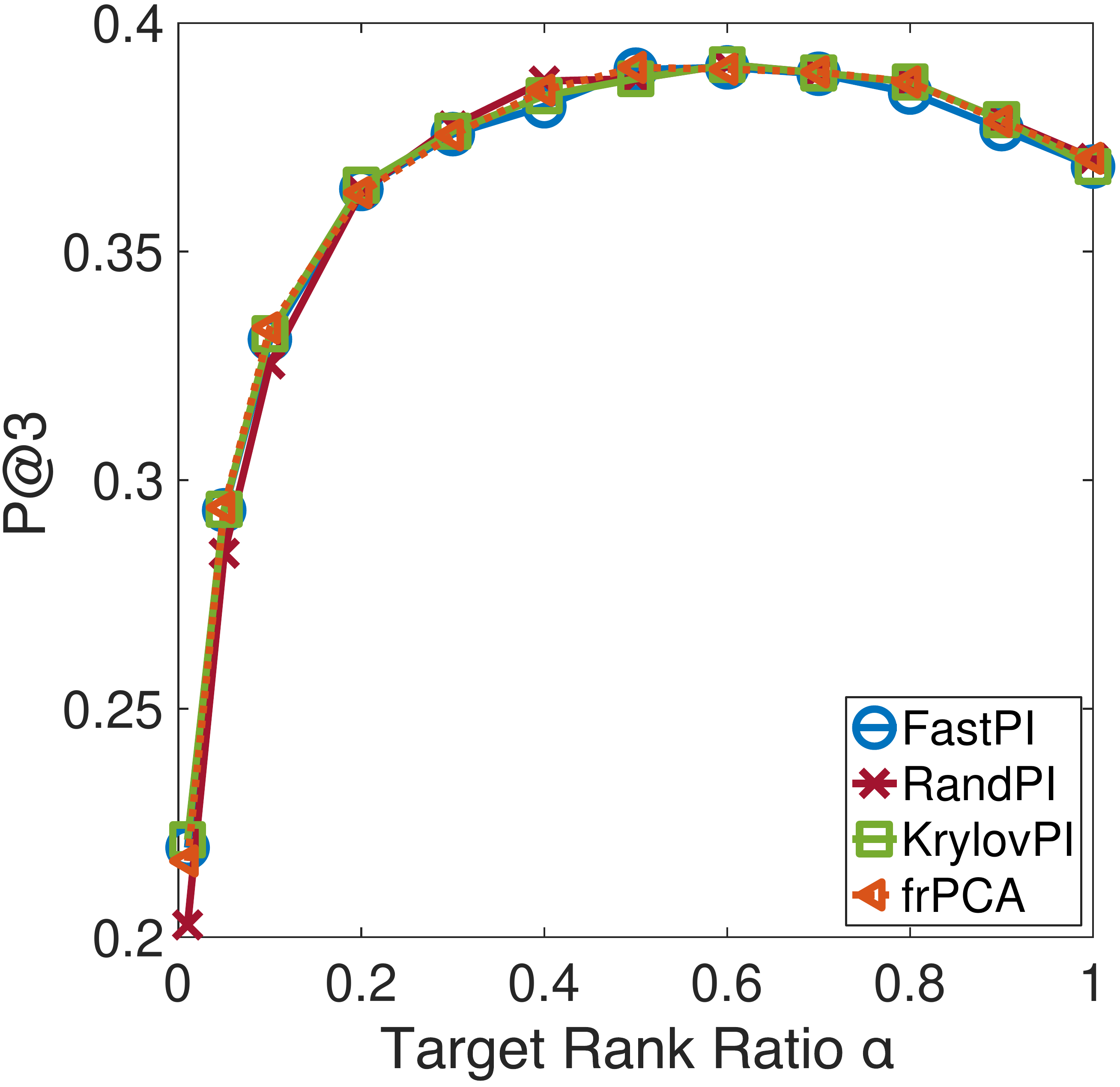}
      }
    \caption{
        \label{fig:accuracy:p3}
        Accuracy of the multi-label linear regression task (Application~\ref{application:mtlr}) in terms of P@3 varying target rank ratio $\alpha$ (Section~\ref{sec:experiment:accuracy}).
        Note that accuracies of all tested methods are almost the same for each $\alpha$, implying that \method accurately computes the approximate pseudoinverse as other methods.
    }
    \vspace{-3mm}
\end{figure*}


\subsection{Accuracy of Multi-label Linear Regression}
\label{sec:experiment:accuracy}
We examine the quality of the approximate pseudoinverse by measuring the predictive performance of each method in the multi-label linear regression task as described in the Application~\ref{application:mtlr}.
For each experiment, we randomly split a multi-label dataset into a training set ($90\%$) and a test set ($10\%$); and compute the pseudoinverse on the training set varying the target rank ratio $\alpha$ from $0.01$ to $1.0$.
Note that in the multi-label datasets, each instance has only a few positive ($1$) labels,  i.e., the label matrix $\mat{Y}$ is sparse as shown in Table~\ref{tab:dataset}.
Therefore, it is important to focus on the accurate prediction of the few positive labels.
Due to this reason, many researchers~\citep{chen2012feature,prabhu2014fastxml,yu2014large} have evaluated the performance of this task using ranking based measures such as top-$k$ precision, denoted by P@$k$, based on predicted scores.
We also measure P@$k$ as the accuracy of this task on the test set.
Let $\vect{y} \in \{0, 1\}^{L}$ be a ground truth label vector and $\vect{\hat{y}} \in \mathbb{R}^{L}$ be its predictive score vector, where $L$ is the number of labels.
Then, $\text{P}@k = \frac{1}{k}\sum_{l \in \text{rank}_k(\vect{\hat{y}})}\vect{y}_{l}$,
where $\text{rank}_k(\vect{\hat{y}})$ returns the $k$ largest indices of $\vect{\hat{y}}$ ranked in the descending order.

Figure~\ref{fig:accuracy:p3} demonstrates the accuracy of each method for the task in terms of P@$3$. 
As expected, the accuracy depends on the rank ratio and there is little difference between the compared methods, which serves to verify our proposed \method is correctly derived.
However, an interesting observation is made about the appropriate rank ratio for uses in multi-linear regression tasks: accuracy plots are curved indicating that there are underfittings when $\alpha$ are too small, and overfittings when $\alpha$ are too large.
This implies that such low-rank approximation with a relatively large rank is effective in the viewpoint of the machine learning application. 


\subsection{Computational Performance}
\label{sec:experiment:computational}
We investigate the computational performance of each method in terms of running time.
For each experiment, we measure the wall-clock time of \method, RandPI,  and KrylovPI varying the target rank ratio $\alpha$ from $0.01$ to $1.0$.

\begin{figure*}[!t]
	\vspace{-2mm}
    \centering
    \subfigure[Amazon]{
        \hspace{-4mm}
        \includegraphics[width=0.233\linewidth]{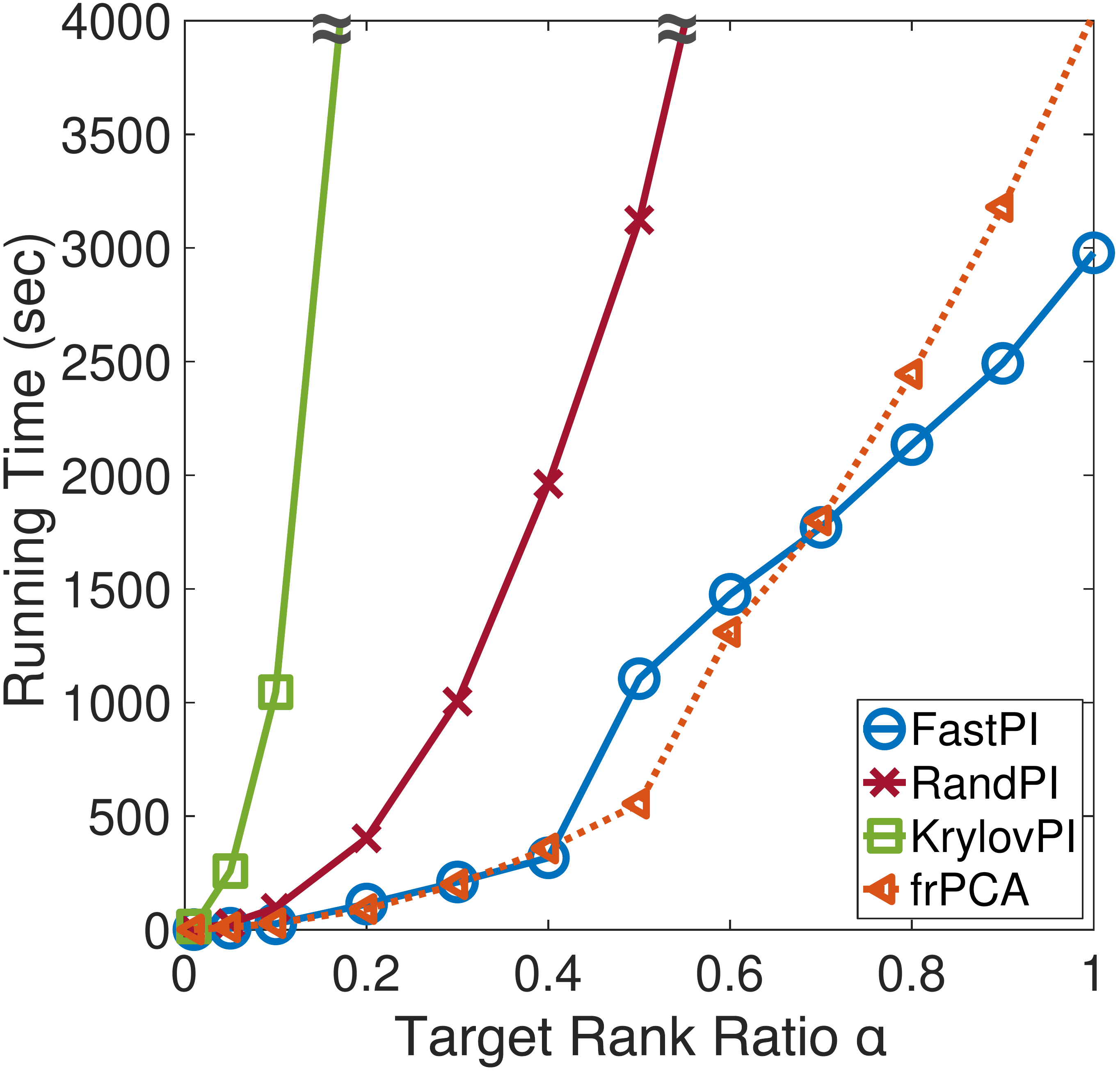}
    }
    \subfigure[RCV]{
        \hspace{-2mm}
        \includegraphics[width=0.228\linewidth]{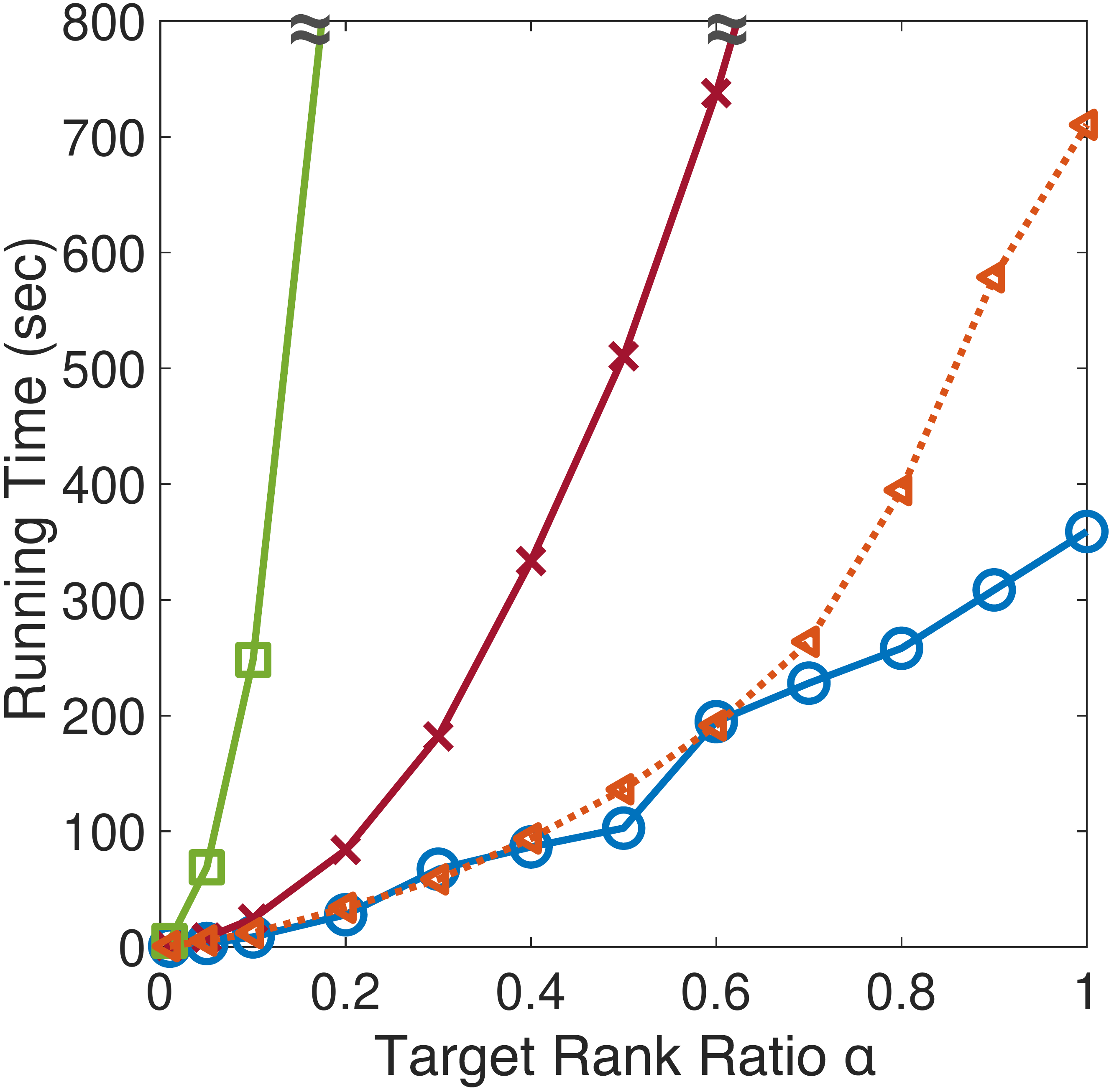}
      }
      \subfigure[Eurlex]{
        \hspace{-2mm}
        \includegraphics[width=0.228\linewidth]{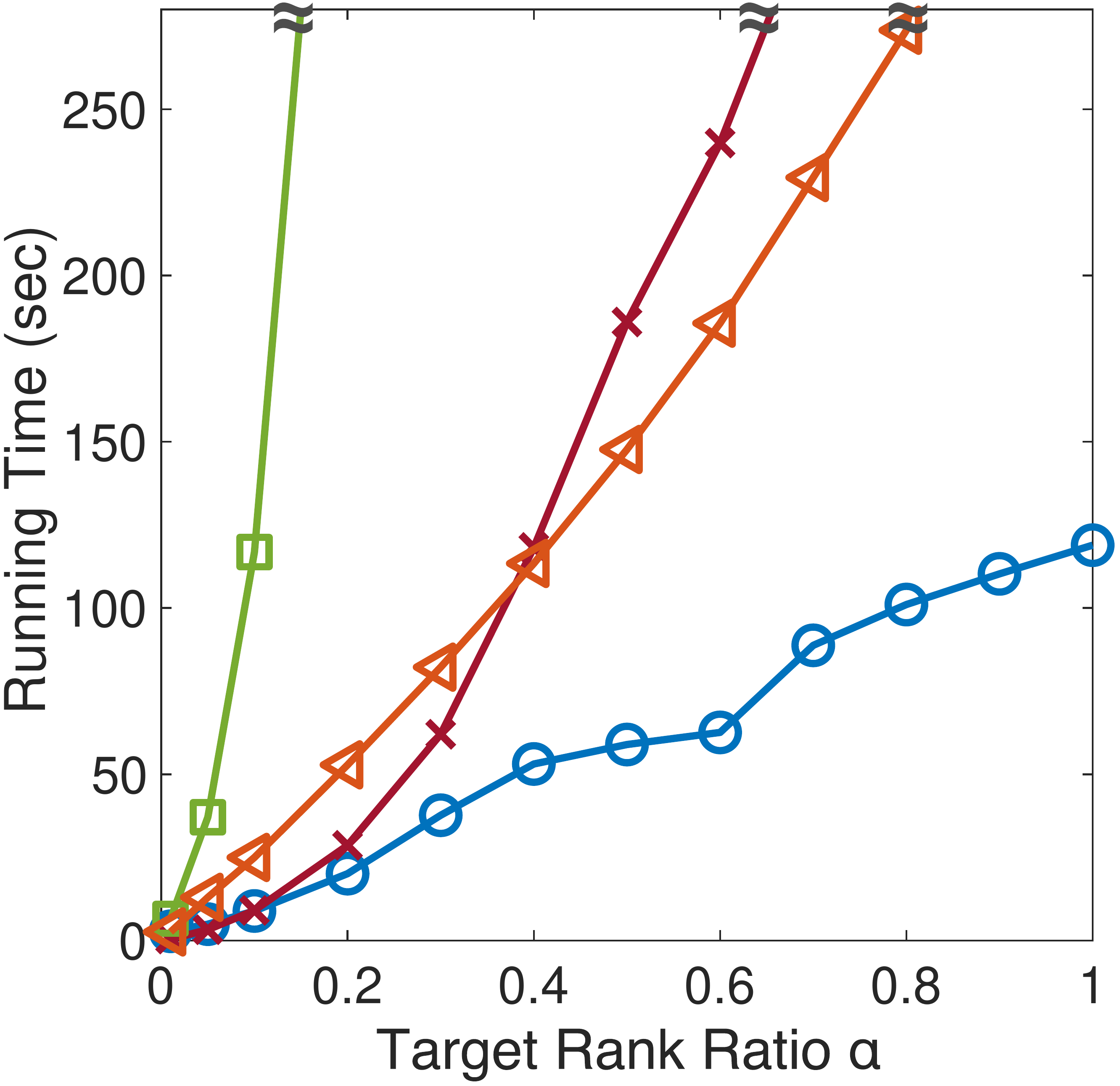}
    }
    \subfigure[Bibtex]{
        \hspace{-2mm}
        \includegraphics[width=0.222\linewidth]{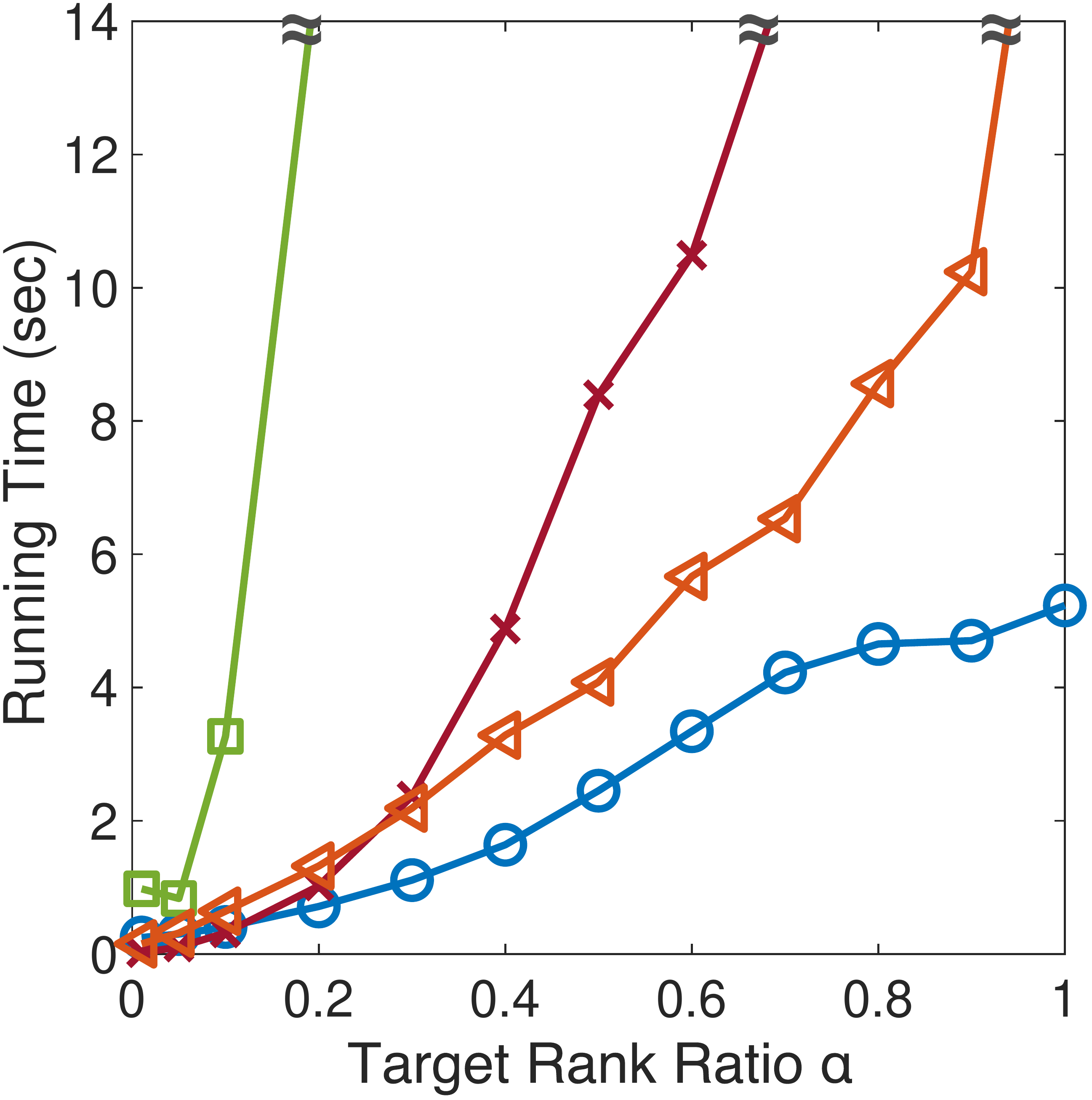}
      }
    \caption{
        \label{fig:running_time}
        Computational performance in terms of running time varying target rank ratio $\alpha$ (Section~\ref{sec:experiment:computational}).
    }
    \vspace{-4mm}
\end{figure*}

Figure~\ref{fig:running_time} demonstrates the running time of methods on each dataset.
First, the running time of KrylovPI, an iterative method, skyrockets since it requires more iterations for convergence as $\alpha$ increases.
KrylovPI is specialized for computing a very few largest or smallest singular values and vectors on a large and sparse matrix (e.g., $\alpha = 0.01$).
Thus, it is impractical to use KrylovPI for obtaining relatively high-rank SVD results as shown in Figure~\ref{fig:running_time}.

We next compare \method to RandPI. 
As shown in Figure~\ref{fig:running_time}, \method is faster than RandPI over all datasets.
Especially, the performance of RandPI becomes much worse as the target rank $r = \lceil \alpha n \rceil$ increases. 
The main reason is because of the oversampling technique of RandPI, i.e., it needs to perform several matrix operations on $m \times 2r$ matrices (see the details in Section~\ref{sec:experiment:setting}).
If $r$ is very small, their computations are efficient.
However, if $r$ is large, and it is close to $n$, then RandPI needs to handle up to $m \times 2n$ matrices whose size is twice the size of original, thereby slowing down the execution speed for large $\alpha$ or high rank.

Finally, we look into the comparison between \method and frPCA. 
Overall, the performance of \method is better than that of frPCA, especially in the Eurlex and Bibtex datasets. 
In the Amazon and RCV datasets, although the running time of \method is similar to or slightly slower than frPCA for $r \leq 0.6$, \method is faster than frPCA for $r > 0.6$.
These results indicate that \method is competitive with frPCA for low ranks, and it is more efficient than frPCA for high ranks.


%% file: 005conclusion.tex
In this paper, we have shown how feature matrix reordering can speed up the approximate pseudoinverse calculation faster than the state-of-the-art low-rank approximation method for computing pseudoinverses.
Our proposed approach \method (\methodlong) is based on a crucial observation that many real-world feature matrices are considerably sparse and skewed, which have the possibility of being reordered. 
\method reorders the feature matrix such that the reordered matrix contains a large and sparse block diagonal matrix whose SVD is easily computed.
\method then efficiently computes the approximate pseudoinverse through incremental low-rank SVD updates. 
We applied the \method on multi-label linear regression problem, a type of machine learning problem. 
Our experiments demonstrate that our \method computes SVD based pseudoinverse quickly for sufficiently large ranks compared to other methods. 